\documentclass[sigconf]{acmart}

\usepackage{hyperref}
\usepackage{url}
\usepackage{booktabs}       
\usepackage{amsfonts}       
\usepackage{nicefrac}       
\usepackage{microtype}      
\usepackage{xcolor}         
\usepackage{colortbl}
\usepackage{graphicx}
\usepackage{multirow}
\usepackage{pifont}
\usepackage{makecell}
\usepackage{listings}
\usepackage{tcolorbox}
\usepackage{rotating}
\usepackage{wrapfig}
\usepackage[normalem]{ulem}
\useunder{\uline}{\ul}{}

\definecolor{ivory2}{RGB}{212,212,212}
\newcommand{\cmark}{\textcolor{green!60!black}{\ding{51}}}  
\newcommand{\xmark}{\textcolor{red}{\ding{55}}}             

\definecolor{codegreen}{rgb}{0,0.6,0}
\definecolor{codegray}{rgb}{0.5,0.5,0.5}
\definecolor{codepurple}{rgb}{0.58,0,0.82}
\definecolor{backcolour}{rgb}{0.95,0.95,0.92}

\lstdefinestyle{mystyle}{
    backgroundcolor=\color{backcolour},   
    commentstyle=\color{codegreen},
    keywordstyle=\color{magenta},
    numberstyle=\tiny\color{codegray},
    stringstyle=\color{codepurple},
    basicstyle=\ttfamily\footnotesize,
    breakatwhitespace=false,         
    breaklines=true,                 
    captionpos=b,                    
    keepspaces=true,                 
    numbers=left,                    
    numbersep=5pt,                  
    showspaces=false,                
    showstringspaces=false,
    showtabs=false,                  
    tabsize=2
}

\AtBeginDocument{%
  }

\copyrightyear{2026}
\acmYear{2026}
\setcopyright{cc}
\setcctype{by}
\acmConference[SIGSPATIAL '26]{The 34th ACM International Conference on Advances in Geographic Information Systems}{November 03--06, 2026}{Riverside, CA, USA}
\acmBooktitle{The 34th ACM International Conference on Advances in Geographic Information Systems (SIGSPATIAL '26), November 03--06, 2026, Riverside, CA, USA}
\acmDOI{10.1145/3841645.3844196}
\acmISBN{979-8-4007-2950-8/2026/11}

\begin{document}

\title{Massive-STEPS: Massive Semantic Trajectories for Understanding POI Check-ins}


\author{Wilson Wongso}
\orcid{0000-0003-0896-1941}
\affiliation{%
    \institution{University of New South Wales}
    \city{Sydney}
    \country{Australia}
}
\email{w.wongso@unsw.edu.au}

\author{Hao Xue}
\orcid{0000-0003-1700-9215}
\affiliation{%
    \institution{University of New South Wales}
    \city{Sydney}
    \country{Australia}
}
\affiliation{%
    \institution{Hong Kong University of Science and Technology (Guangzhou)}
    \city{Guangzhou}
    \country{China}
}
\email{haoxue@hkust-gz.edu.cn}

\author{Flora D. Salim}
\orcid{0000-0002-1237-1664}
\affiliation{%
    \institution{University of New South Wales}
    \city{Sydney}
    \country{Australia}
}
\email{flora.salim@unsw.edu.au}

\renewcommand{\shortauthors}{Wongso et al.}

\begin{abstract}
    Understanding human mobility through Point-of-Interest (POI) trajectory modeling is increasingly important for applications such as urban planning, personalized services, and generative agent simulation. However, progress in this field is hindered by two key challenges: the over-reliance on older datasets from 2012-2013 and the lack of reproducible, city-level check-in datasets that reflect diverse global regions. To address these gaps, we present Massive-STEPS (Massive Semantic Trajectories for Understanding POI Check-ins), a large-scale, publicly available benchmark dataset built upon the Semantic Trails dataset and enriched with semantic POI metadata. Massive-STEPS spans 15 geographically and culturally diverse cities and features check-in data from both 2012-2013 and 2017-2018, with a longer duration of 24 months than prior datasets. We benchmarked a wide range of POI models on Massive-STEPS using both supervised and zero-shot approaches, and evaluated their performance across multiple urban contexts. By releasing Massive-STEPS, we aim to facilitate reproducible and equitable research in human mobility and POI trajectory modeling. The dataset and benchmarking code are available at: \url{https://github.com/cruiseresearchgroup/Massive-STEPS}.
\end{abstract}

\ccsdesc[500]{Information systems~Data mining}
\ccsdesc[300]{Information systems~Recommender systems}
\ccsdesc[300]{Information systems~Spatial-temporal systems}

\keywords{Human Mobility, Point-of-Interest Recommendation, Benchmark}


\maketitle

\section{Introduction}
\label{sec:introduction}

\paragraph{Importance of Human Mobility Data and Modeling}

Human mobility data and Point-of-Interest (POI) trajectory modeling are essential for understanding how individuals interact with and move through physical spaces. It enables applications such as urban planning \cite{yuan2025worldmove}, travel service recommendations \cite{feng-etal-2025-agentmove}, commercial advertising \cite{10.1145/3477495.3531983}, and POI recommendation \cite{10.1145/3369822,10.1145/3626772.3657840,zhang2025survey}. Human mobility data has become even more crucial with the increasing use of large language model (LLM) agents to simulate human-like behavior and routines~\cite{zhou2024large,jiawei2024large}. However, while simulated and aggregated human mobility data are starting to gain popularity \cite{10.1145/3394486.3412862,10.1145/3624475,10.1145/3681765.3698455,Jiang03072025}, they may not accurately reflect real-world, fine-grained individual human behavior \cite{SALIM2020106964}, highlighting the value of real-world data. These advancements are enabled by and large with Location-based Social Networks (LBSNs), which generate vast amounts of spatio-temporal data through user check-ins \cite{zhang2025survey,10.1145/3626772.3657840}. This rich data source has allowed the development of POI recommendation systems that leverage users' historical visits to recommend relevant locations. Such systems enhanceuser engagement and commercial value through personalized recommendations \cite{10.1145/3369822}.

\paragraph{Literature Gaps}

\begin{figure*}[tbp]
    \centering
    \includegraphics[width=0.92\textwidth]{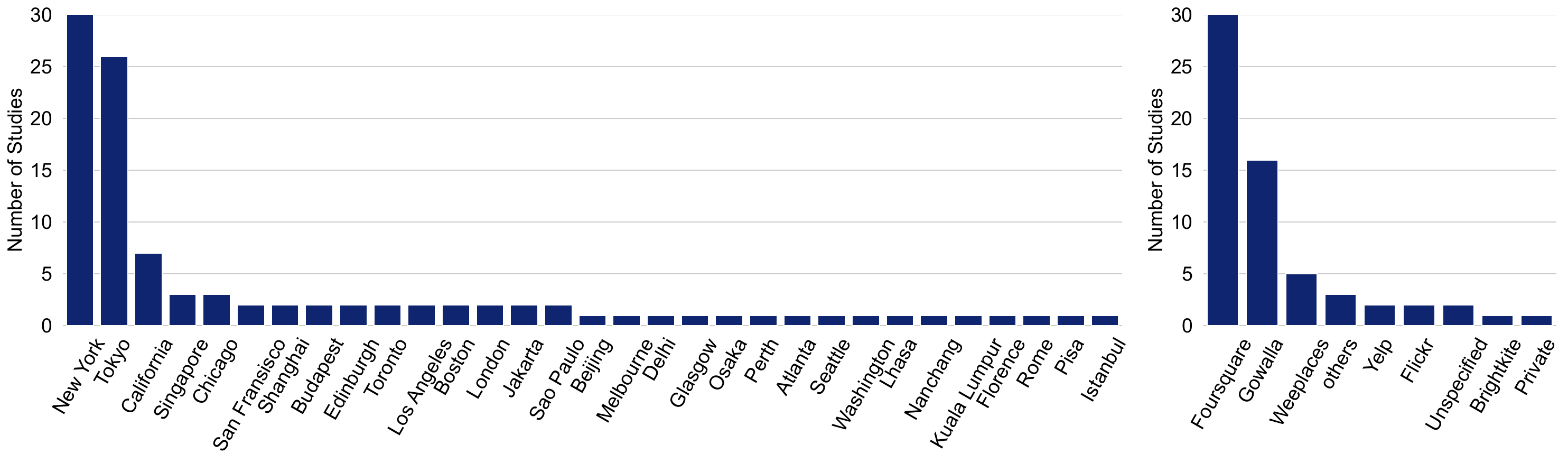}
    \caption{\textbf{Distribution of POI recommendation studies} modeled on specific cities, modified from Table IV of \cite{zhang2025survey}. We identified and counted studies that explicitly mentioned city names, revealing the skewness of existing research, which is saturated around New York and Tokyo, sourced primarily from Foursquare.}
    \label{fig:city-studies}
    \Description{Distribution of POI recommendation studies modeled on specific cities, modified from Table IV of \cite{zhang2025survey}. We identified and counted studies that explicitly mentioned city names, revealing the skewness of existing research, which is saturated around New York and Tokyo, sourced primarily from Foursquare.}
\end{figure*}

Our paper addresses three critical gaps in POI trajectory modeling research and datasets. First, as shown in Fig.~\ref{fig:city-studies}, the field remains dominated by studies focused on just two cities, New York and Tokyo, based on the Foursquare dataset curated by \citet{yang2014modeling}. This dataset, collected in 2012-2013, raises concerns about its temporal quality, as many POIs may no longer exist and user behavior may have changed \cite{yeow2021point}. While some recent studies have expanded to other cities \cite{10.1007/s11280-024-01279-y,merinov2024positive,feng-etal-2025-agentmove}, they often rely on the Global-scale Check-in Dataset (GSCD) \cite{yang2015nationtelescope,yang2016participatory}, which, despite its extensive coverage, is also from 2012-2013 and contains nearly 50\% erroneous entries \cite{monti2018semantic}. Second, most existing studies are difficult to reproduce, either due to the lack of clearly defined geographic boundaries or the unavailability of the datasets themselves, hindering fair comparison and replication. Finally, we join recent efforts~\cite{yuan2025worldmove} in advocating for the inclusion of low-resource and underrepresented cities. Expanding beyond well-studied urban centers is essential for building more generalizable and universally applicable POI models. Table~\ref{tab:dataset-comparison} summarizes these limitations in terms of geographic coverage, temporal span, and reproducibility.

\begin{figure*}[tbp]
    \centering
    \includegraphics[width=0.925\textwidth]{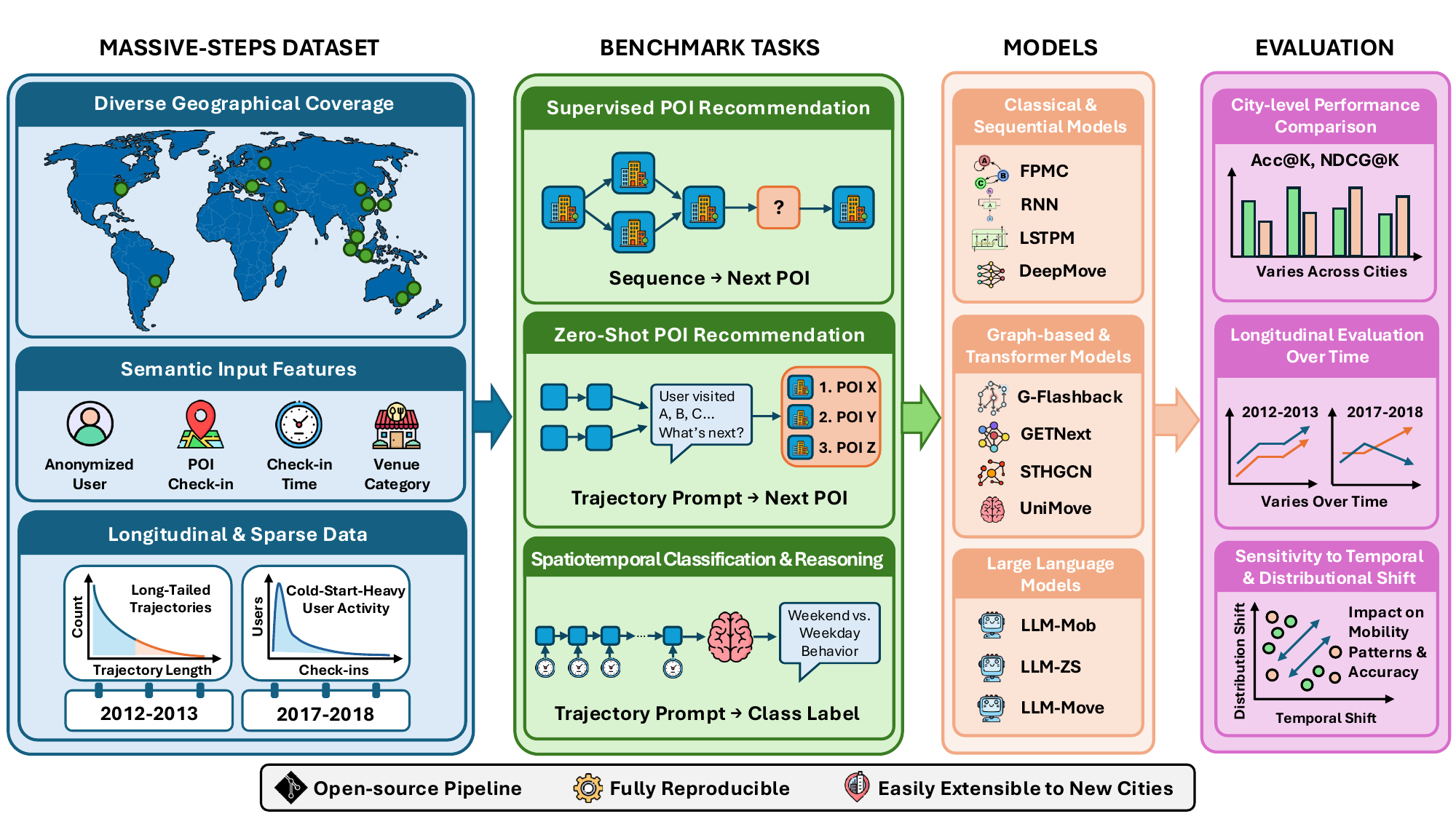}
    \caption{\textbf{Massive-STEPS Dataset and Benchmark Overview.}}
    \label{fig:all-tasks}
    \Description{Massive-STEPS Dataset and Benchmark Overview.}
\end{figure*}

\paragraph{Massive-STEPS Dataset}

In this paper, we introduced the Massive Semantic Trajectories for Understanding POI Check-ins (Massive-STEPS) Dataset, derived from the Semantic Trails dataset (STD) \cite{monti2018semantic}. Massive-STEPS includes check-ins from 2012-2013 and 2017-2018, providing updated POI check-in data. This supports longitudinal POI trajectory modeling studies and addresses the limitations of older datasets commonly used in prior studies. The dataset covers 15 diverse cities spanning East, Southeast, and Western Asia, Europe, North and South America, and Australia. Notably, we placed a deliberate emphasis on under-explored regions by including cities such as Jakarta, Kuwait City, and Petaling Jaya, filling a key gap in POI trajectory research that has largely focused on major urban centers. We further enriched STD by aligning it with Foursquare's Open Source Places dataset, incorporating missing metadata such as geographic coordinates, POI names, and addresses.

\paragraph{Benchmark Tasks}
To demonstrate the utility of this dataset, we conducted an extensive benchmark on three tasks as shown in Fig.~\ref{fig:all-tasks}: (1) supervised POI recommendation, (2) zero-shot POI recommendation, and (3) spatiotemporal classification and reasoning. Our benchmark covers a wide range of models, including traditional approaches, deep learning-based models, and more recent LLM-based methods. The goal of POI recommendation task is to predict a set of POIs that a user is likely to visit based on their current check-in trajectory and historical behavior. This reflects real-world applications such as personalized POI recommendations in location-based services. Similarly, the goal of spatiotemporal classification and reasoning is to assess how effectively models (e.g., LLMs) leverage, interpret, and reason about POI trajectories. In addition, the scale of our dataset allows us to examine how urban features influence POI modeling accuracy. Building on prior hypotheses, we propose a new insight: cities with more evenly distributed POI categories tend to be harder to model, as the absence of a dominant POI category makes user behavior less predictable.

\begin{table}[tbp]
    \centering
    \scriptsize
    \caption{\textbf{Comparison of check-in datasets commonly used for POI modeling tasks}.}
    \label{tab:dataset-comparison}
    \setlength{\tabcolsep}{1.5pt}
    \begin{tabular}{lccc c cc}
        \toprule
        \multirow{2}{*}{\textbf{Dataset}}                                 &
        \multicolumn{3}{c}{\textbf{Scale}}                                &
        \multicolumn{1}{c}{\textbf{\phantom{xx}Completeness\phantom{xx}}} &
        \multicolumn{2}{c}{\textbf{Usability}}                                                                                                                                                                                                                                                                                \\
        \cmidrule(lr){2-4} \cmidrule(lr){5-5} \cmidrule(lr){6-7}
                                                                          & \textbf{\#cities}            & \textbf{Years}                                         & \textbf{\#months}            &
        \textbf{POI Attributes}                                           & \textbf{Replicable}          & \textbf{Open}                                                                                                                                                                                                      \\
        \midrule
        NYC/TKY \cite{yang2014modeling}                                   & 2                            & 2012-2013                                              & 11                           & \hspace{6pt}Coordinates, Category                                      & \cmark                  & \cmark                  \\
        Gowalla-CA \cite{10.1145/2484028.2484030}                         & 1                            & 2009-2010                                              & 21                           & \hspace{6pt}Coordinates, Category                                      & \cmark                  & \cmark                  \\
        AgentMove \cite{feng-etal-2025-agentmove}                         & 12                           & 2012-2013                                              & 17                           & \hspace{6pt}Coordinates, Category                                      & \xmark                  & \xmark                  \\
        \midrule
        \multirow{2}{*}{\textbf{Massive-STEPS}}                           & \multirow{2}{*}{\textbf{15}} & \makecell[tc]{\textbf{2012-2013,}\\\textbf{2017-2018}} & \multirow{2}{*}{\textbf{24}} & \makecell[tc]{\textbf{Coordinates, Category,}\\\textbf{Name, Address}} & \multirow{2}{*}{\cmark} & \multirow{2}{*}{\cmark} \\
        \bottomrule
    \end{tabular}
\end{table}

\paragraph{Contribution}

This paper introduces the Massive Semantic Trajectories for Understanding POI Check-ins (Massive-STEPS) dataset, addressing gaps in existing POI trajectory modeling research. Current POI check-in datasets are often only from 2012-2013, skewed to a few cities, and lack semantic metadata, hindering the development of robust and globally applicable models. While datasets like GSCD and STD offer broad geographic coverage, they either suffer from an older timespan, contain erroneous data, or have missing information. Massive-STEPS addresses these issues by providing high-quality check-ins from 2012-2013 and 2017-2018, allowing for longitudinal POI trajectory modeling studies. The dataset spans 15 diverse cities across multiple regions, with a focus on low-resource cities overlooked in previous research. Additionally, Massive-STEPS is enriched with metadata through alignment with Foursquare's Open Source Places, providing crucial details such as POI geographical coordinates, POI names, and addresses. We also conducted an extensive benchmark on both supervised and zero-shot POI recommendation and trajectory classification tasks, evaluating a wide range of models, including traditional methods, deep learning approaches, and recent LLM-based techniques. We further analyzed which urban features affect POI modeling accuracy and found that cities with no dominant POI category tend to be harder to predict. By releasing this dataset and benchmark code publicly, we facilitate open and reproducible research, enabling future advancements in urban mobility studies.

\section{Related Works}

\subsection{Existing Datasets}
\label{sec:existing-datasets}

A survey conducted by \cite{zhang2025survey} outlines the landscape of POI trajectory modeling research, covering a wide range of models and architectures used in prior studies. While it offers a high-level overview of the datasets used, it lacks a dedicated discussion or evaluation of POI datasets. We address this gap by analyzing commonly used datasets and positioning our dataset within this context.

\paragraph{LBSN Check-in Data Sources}

Building on the table from \cite{zhang2025survey}, we investigated which datasets are most commonly used in prior studies. From their original table (Table IV), we filtered entries pertaining specifically to POI and next POI recommendation tasks and identified (1) the most frequently used LBSN check-in data sources and (2) the most commonly studied cities. As shown in Fig.~\ref{fig:city-studies}, Foursquare remains the dominant source of LBSN data in existing studies, appearing in almost 50\% of the surveyed works. While several variants of Foursquare datasets have been employed, the most widely used are the NYC and Tokyo Dataset \cite{yang2014modeling} (often abbreviated as FSQ-NYC and FSQ-TKY) and the Global-scale Check-in Dataset (GSCD) \cite{yang2015nationtelescope,yang2016participatory}, curated by the same authors. Other LBSN sources occasionally used include Gowalla \cite{cho2011friendship}, Brightkite \cite{cho2011friendship}, and Weeplaces \cite{10.14778/3115404.3115407}.

\begin{table}[tbp]
    \centering
    \footnotesize
    \caption{\textbf{Overview of POI dynamics}: total POIs ever opened (according to Foursquare OS Places), POIs confirmed closed by 2025, and POIs closed during 2014-2016, corresponding to the temporal gap in our dataset.}
    \label{tab:poi-closures}
    \setlength{\tabcolsep}{3pt}
    \begin{tabular}{lrrr}
        \toprule
        \textbf{City} & \multicolumn{1}{c}{\textbf{POIs Opened}} & \multicolumn{1}{c}{\textbf{Total POIs Closed}} & \multicolumn{1}{c}{\textbf{Closed within 2014-2016}} \\
        \midrule
        New York      & 49,218                                   & 13,009 (26.43\%)                               & 3,118 (6.34\%)                                       \\
        Melbourne     & 7,699                                    & 1,850 (24.03\%)                                & 209 (2.71\%)                                         \\
        Sydney        & 8,986                                    & 1,759 (19.57\%)                                & 253 (2.82\%)                                         \\
        Moscow        & 17,822                                   & 3,021 (16.95\%)                                & 868 (4.87\%)                                         \\
        São Paulo     & 38,377                                   & 4,990 (13.00\%)                                & 1,257 (3.28\%)                                       \\
        Shanghai      & 4,462                                    & 661 (14.81\%)                                  & 81 (1.82\%)                                          \\
        Tokyo         & 4,725                                    & 421 (8.91\%)                                   & 0 (0.00\%)                                           \\
        Petaling Jaya & 60,158                                   & 4,186 (6.96\%)                                 & 1,533 (2.55\%)                                       \\
        Istanbul      & 53,812                                   & 2,833 (5.26\%)                                 & 481 (0.89\%)                                         \\
        Beijing       & 1,127                                    & 56 (4.97\%)                                    & 10 (0.89\%)                                          \\
        Jakarta       & 76,116                                   & 3,527 (4.63\%)                                 & 483 (0.63\%)                                         \\
        Bandung       & 29,026                                   & 1,053 (3.63\%)                                 & 182 (0.63\%)                                         \\
        Palembang     & 4,343                                    & 143 (3.29\%)                                   & 23 (0.53\%)                                          \\
        Tangerang     & 12,956                                   & 383 (2.96\%)                                   & 50 (0.39\%)                                          \\
        Kuwait City   & 17,180                                   & 161 (0.94\%)                                   & 22 (0.13\%)                                          \\
        \bottomrule
    \end{tabular}
\end{table}

\paragraph{Saturated to Two Cities and Old Timespan}

Due to the widespread use of FSQ-NYC and FSQ-TKY \cite{yang2014modeling}, the majority of POI trajectory studies are disproportionately focused on these two cities, as illustrated in Fig.~\ref{fig:city-studies}. While there is nothing inherently problematic about NYC and Tokyo, there has been growing interest in expanding research to a broader range of cities, particularly those that are underexplored or considered low-resource \cite{yuan2025worldmove}, as cultural and regional differences influence collective mobility behaviors. For instance, in some cities, residents tend to commute to business districts in the morning, whereas in others, nightlife activities such as visiting bars after work are more common \cite{yang2015nationtelescope}. Ensuring diverse geographic coverage is increasingly important, especially as LLMs are adopted for POI trajectory modeling tasks and are known to exhibit geographical biases \cite{manvi2024large}.

\begin{figure}[bp]
    \centering
    \includegraphics[width=0.475\textwidth]{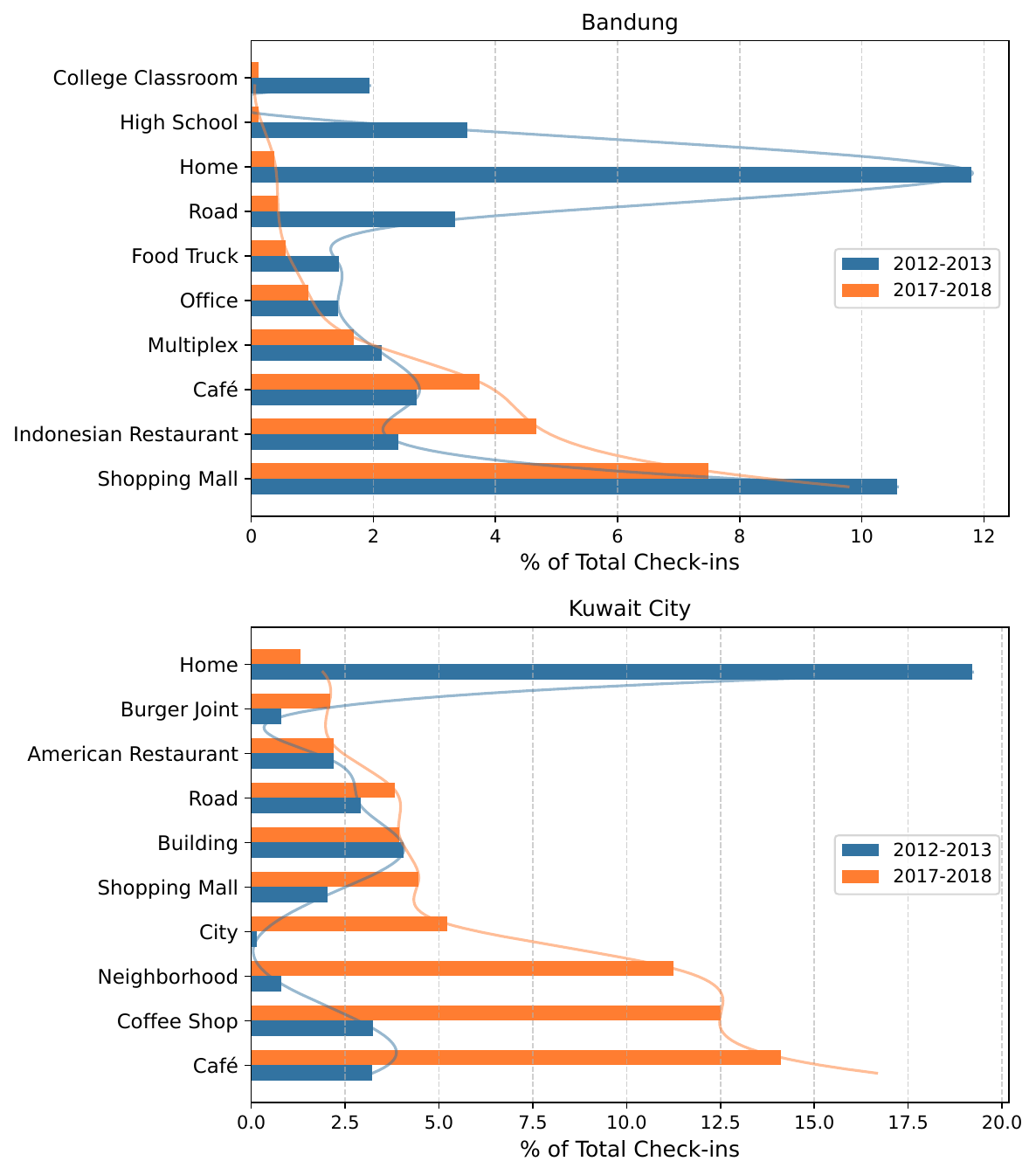}
    \caption{Top 10 most visited POI categories in Bandung and Kuwait City across two time periods.
    }
    \label{fig:top-poi-categories-small}
    \Description{Top 10 most visited POI categories in Bandung and Kuwait City across two time periods.}
\end{figure}

In addition, because many studies rely on the FSQ-NYC and FSQ-TKY, they are often constrained to the timespan it covers: check-in data from 2012 to 2013. However, POI data is inherently dynamic: venues may have closed, relocated, or changed in category over time. Table~\ref{tab:poi-closures} illustrates this by detailing the total number of POIs ever opened, the number of POIs closed by 2025, and the number of POIs that closed between 2014 and 2016 alone. These substantial closure rates in major cities highlight the critical need for newer, multi-period datasets that reflect the actual lifecycle of POIs \cite{yeow2021point}. This is particularly critical, as recommender systems should avoid suggesting POIs that no longer exist or have undergone substantial changes (e.g., a former bookstore converted into a coworking space) and behave dynamically over longitudinal periods \cite{Yabe2024}. Moreover, visitation patterns may change over time depending on the city (see Fig.~\ref{fig:top-poi-categories-small}), potentially due to shifts in routines. For instance, the opening of a new train station can noticeably alter commuting patterns and the popularity of nearby POIs.

\paragraph{Low Data Quality: Erroneous Entries}

More recently, researchers have begun leveraging the broader Global-scale Check-in Dataset (GSCD) \cite{yang2015nationtelescope,yang2016participatory}, which spans 415 cities across 77 countries. Despite its wider geographic coverage, GSCD is temporally limited to the same 2012-2013 period as FSQ-NYC and FSQ-TKY, and thus suffers from similar issues of temporal quality. More critically, \citet{monti2018semantic} demonstrated that GSCD suffers from significant data quality issues, with over 14 million check-ins (about 44\%) of the dataset flagged as erroneous due to anomalous user behavior. These include (1) repeated check-ins at the same venue, (2) check-ins occurring within implausibly short time intervals (less than one minute), and (3) transitions between venues that would require travel speeds exceeding Mach 1, which are physically impossible.

To address these limitations, \citet{monti2018semantic} introduced the Semantic Trails Dataset (STD), which applies systematic filtering procedures to enhance data quality. STD comprises two subsets: a cleaned version of GSCD covering 2012-2013 (STD 2013), and a newer collection of check-ins from 2017-2018 (STD 2018), sourced from Foursquare Swarm. STD 2018 also spans a wider range of cities, making it valuable for capturing globally distributed user behavior, in contrast to GSCD's focus on densely populated urban centers. Both subsets follow the same rigorous filtering criteria, resulting in a higher-quality check-in dataset for downstream POI trajectory modeling tasks. Given these improvements, we adopted STD as the source for our check-in dataset.

\paragraph{Poor Reproducibility}

Another persistent challenge in POI trajectory research is the lack of reproducibility in dataset preprocessing. While some recent studies utilize datasets like GSCD to cover a wide range of cities, they often omit important details needed for replicating their data filtering processes. For example, \cite{feng-etal-2025-agentmove} and \cite{zuo2024diff} conducted city-level filtering, but they did not specify how the city boundaries were defined. Similarly, the Weeplaces dataset used in \cite{chen2025self} and \cite{cao2023improving} is no longer available. Examining all studies reviewed in \cite{zhang2025survey}, we found that \textbf{almost none of the datasets used in these works are fully reproducible or publicly available}, except for FSQ-NYC/TKY \cite{yang2014modeling} and Gowalla-CA \cite{10.1145/2484028.2484030}, explaining the heavy reliance on these datasets.


\subsection{Understanding Urban Features and POI Trajectory Modeling}
\label{sec:understanding-urban-features}

POI trajectory studies that evaluate models across multiple city-level datasets often include analyses to assess how well their methods generalize across different urban contexts. It is well understood that POI recommendation accuracy metrics (e.g., Acc@k, NDCG@k) can vary substantially between cities and can be interpreted as a proxy for how easy or difficult a city is to model. The assumption is that higher performance reflects more predictable or structured mobility patterns. This viewpoint is consistent with prior work highlighting the role of cultural and urban-specific factors in shaping mobility behaviors \cite{yang2015nationtelescope,sun2024city}. For example, Gowalla-CA \cite{cho2011friendship,10.1145/2484028.2484030} often yields lower accuracy compared to FSQ-NYC and FSQ-TKY \cite{yang2014modeling}, suggesting that some cities may be inherently harder to model.

Several studies have proposed hypotheses connecting specific urban features to modeling difficulty. \citet{10.1145/3477495.3531983} hypothesized that cities with fewer check-ins and higher spatial sparsity of POIs are harder to model. \citet{10.1145/3539618.3591770} suggested that a larger number of user trajectories improves predictive accuracy by providing richer collaborative signals, whose architecture is designed to leverage. \citet{10.1145/3626772.3657840} proposed that cities with a greater variety of POI categories are easier to model due to LLMs' contextual reasoning capabilities, whereas cities covering a broader geographic area tend to be more difficult to model. In the zero-shot POI recommendation setting, \citet{feng-etal-2025-agentmove} reported two key findings: (1) geospatial biases inherent in LLMs can hinder prediction quality across cities, and (2) LLMs are influenced by city-specific mobility patterns.

Building on these insights, we used Massive-STEPS to explore how urban features affect POI recommendations. Its diverse set of 15 cities allows for a comprehensive analysis across different cultural and urban contexts. We analyzed the correlation between urban features and model accuracy, and based on the results, proposed a new hypothesis that contrasts previous findings in the literature.

\section{Massive-STEPS Dataset}

\subsection{Creation Process}

Massive-STEPS is derived from Semantic Trails\footnote{Semantic Trails is originally released with a CC0 1.0 Universal license.} \cite{monti2018semantic}, incorporating check-ins from both the 2013 and 2018 subsets. We utilize two additional components from STD: (1) the \texttt{cities} metadata file, which provides the latitude and longitude of administrative regions along with their corresponding country codes obtained from GeoNames; and (2) the POI \texttt{category} mapping, which links each Foursquare Category ID to its descriptive name. Based on this metadata, each POI is thus associated with several attributes: Foursquare Place ID (e.g. \texttt{4b1ef9a0f964a520602224e3}), Foursquare Category ID (e.g. \texttt{4bf58dd8d48988d1ae941735}), category name (e.g. \texttt{University}), latitude/longitude of the administrative region (e.g. \texttt{-33.92399,\allowbreak 151.22749}), the administrative region name (e.g. \texttt{Kingsford}), and the country code (e.g. \texttt{AU}). For privacy reasons and model training purposes, we apply ordinal encoding to Place and Category IDs, assigning each a unique integer index.

\paragraph{Trajectory Grouping}
\label{sec:data-preprocessing}

Most POI trajectory models operate on sequences of check-ins, commonly referred to as trajectories. The model is tasked with predicting the next POIs a user is likely to visit, given the current trajectory. STD conveniently provides pre-grouped trajectories (trails) by applying a time interval-based grouping: for each user, check-ins that occur within a time interval of $\delta_\tau = 8$ hours are grouped into the same trajectory. 

\begin{table}[tbp]
    \centering
    \scriptsize
    \caption{\textbf{Statistics} of 15 Massive-STEPS subsets, including the number of users, trajectories, POIs, check-ins, and train, validation, and test trajectory counts. $\mu_{\text{len}}$ denotes the mean number of check-ins per trajectory, and $\mu_{\text{int}}$ denotes the mean time interval between check-ins (in hours). For comparison, we also include statistics from existing Foursquare- and Gowalla-based datasets. $^\dagger$We report the version used in \cite{10.1145/3539618.3591770}.}
    \label{tab:semantic-stats}
    \setlength{\tabcolsep}{2.3pt}
    \begin{tabular}{lrrrrrrrcc}
        \toprule
        \textbf{City} & \textbf{Users} & \textbf{Trajectories} & \textbf{POIs} & \textbf{Check-ins} & \textbf{\#train} & \textbf{\#val} & \textbf{\#test} & \textbf{$\mu_{\text{len}}$} & \textbf{$\mu_{\text{int}}$} \\
        \midrule
        \rowcolor{ivory2}
        \multicolumn{10}{l}{\textbf{NYC and Tokyo Check-in Dataset$^\dagger$ \cite{yang2014modeling}}}                                                                                                                \\
        New York      & 1,048          & 14,130                & 4,981         & 103,941            & 72,206           & 1,400          & 1,347           & 7.55                        & 7.27                        \\
        Tokyo         & 2,282          & 65,499                & 7,833         & 405,000            & 274,597          & 6,868          & 7,038           & 6.32                        & 5.47                        \\
        \midrule
        \rowcolor{ivory2}
        \multicolumn{10}{l}{\textbf{Gowalla$^\dagger$ \cite{cho2011friendship,10.1145/2484028.2484030}}}                                                                                                              \\
        California    & 3,957          & 45,123                & 9,690         & 238,369            & 154,253          & 3,529          & 2,780           & 5.24                        & 8.37                        \\
        \midrule
        \rowcolor{ivory2}
        \multicolumn{10}{l}{\textbf{Massive-STEPS}}                                                                                                                                                                   \\
        Bandung       & 3,377          & 55,333                & 29,026        & 161,284            & 38,732           & 5,534          & 11,067          & 2.91                        & 3.17                        \\
        Beijing       & 56             & 573                   & 1,127         & 1,470              & 400              & 58             & 115             & 2.57                        & 3.10                        \\
        Istanbul      & 23,700         & 216,411               & 53,812        & 544,471            & 151,487          & 21,641         & 43,283          & 2.52                        & 4.36                        \\
        Jakarta       & 8,336          & 137,396               & 76,116        & 412,100            & 96,176           & 13,740         & 27,480          & 3.00                        & 2.81                        \\
        Kuwait City   & 9,628          & 91,658                & 17,180        & 232,706            & 64,160           & 9,166          & 18,332          & 2.54                        & 5.31                        \\
        Melbourne     & 646            & 7,864                 & 7,699         & 22,050             & 5,504            & 787            & 1,573           & 2.80                        & 3.27                        \\
        Moscow        & 3,993          & 39,485                & 17,822        & 105,620            & 27,639           & 3,949          & 7,897           & 2.67                        & 3.36                        \\
        New York      & 6,929          & 92,041                & 49,218        & 272,368            & 64,428           & 9,204          & 18,409          & 2.96                        & 3.16                        \\
        Palembang     & 267            & 4,699                 & 4,343         & 14,467             & 3,289            & 470            & 940             & 3.08                        & 3.17                        \\
        Petaling Jaya & 14,308         & 180,410               & 60,158        & 506,430            & 126,287          & 18,041         & 36,082          & 2.81                        & 2.96                        \\
        São Paulo     & 5,822          & 89,689                & 38,377        & 256,824            & 62,782           & 8,969          & 17,938          & 2.86                        & 3.54                        \\
        Shanghai      & 296            & 3,636                 & 4,462         & 10,491             & 2,544            & 364            & 728             & 2.89                        & 3.02                        \\
        Sydney        & 740            & 10,148                & 8,986         & 29,900             & 7,103            & 1,015          & 2,030           & 2.95                        & 3.33                        \\
        Tangerang     & 1,437          & 15,984                & 12,956        & 45,521             & 11,188           & 1,599          & 3,197           & 2.85                        & 3.24                        \\
        Tokyo         & 764            & 5,482                 & 4,725         & 13,839             & 3,836            & 549            & 1,097           & 2.52                        & 5.16                        \\
        \bottomrule
    \end{tabular}
\end{table}

\paragraph{Matching Trajectories to Target Cities}

To obtain city-specific datasets, we matched trajectories to the target cities. For each city, we obtain geographic boundaries from OpenStreetMap and retrieve its GeoJSON file via the Overpass API. The GeoJSON file contains a polygon defining the city's boundary in latitude and longitude. Using this boundary, we filter check-ins by comparing the latitude/longitude of each POI's administrative region and retain only those that are within the city's polygon. This ensures that all retained trajectories are spatially grounded within the designated city. For reproducibility purposes, we also include the GeoJSON files with our dataset's release.

\paragraph{Filtering Short Trajectories and Inactive Users}

To ensure data quality, we apply an additional filtering step by removing trajectories with fewer than two check-ins and excluding users with fewer than three trajectories. This prevents the model from learning from overly sparse or irrelevant data.

\paragraph{Train, Validation, and Test Splits}

We split trajectories into training, validation, and test sets in a ratio of 7:1:2, following \cite{feng-etal-2025-agentmove}. We ensure that all users in the test set appear at least once in the training or validation set, following prior studies \cite{10.1145/3477495.3531983,10.1145/3539618.3591770}.

\paragraph{POI Enrichment via Foursquare OS Places}
\label{sec:poi-enrichment}

Since the POIs in STD include their corresponding Foursquare Place IDs, we matched them directly with entries in the Foursquare OS Places dataset\footnote{Foursquare OS Places is originally released with an Apache 2.0 license.} using these IDs as the key. This one-to-one ID correspondence allows for a straightforward join operation, enriching each POI with additional metadata such as its latitude and longitude, name (e.g., of a restaurant or subway station), and address. The venue's latitude and longitude are essential for POI recommendation models, as they enable modeling spatial relationships between venues. However, not all POIs in the Foursquare OS Places dataset include the full metadata, particularly those categorized as private residences, which are expectedly excluded due to privacy restrictions\footnote{13.25\% of check-ins do not have exact venue-level coordinates.}.

\begin{figure}[tbp]
    \centering
    \includegraphics[width=0.39\textwidth]{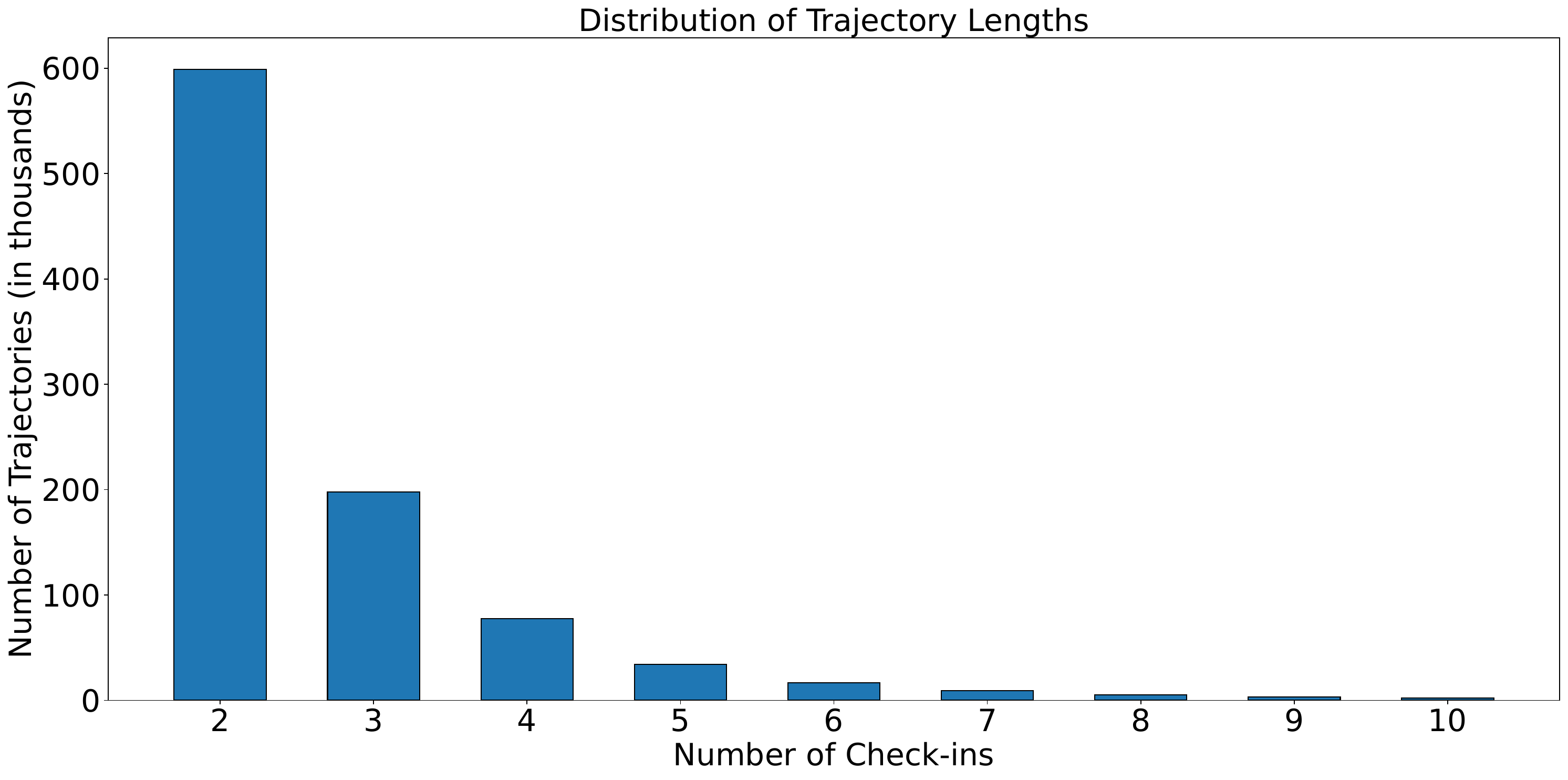}
    \caption{\textbf{Distribution of trajectory lengths}, showing a long-tailed pattern with most trajectories consisting of only 2 to 3 check-ins.}
    \label{fig:trail-length-distribution}
    \Description{Distribution of trajectory lengths, showing a long-tailed pattern with most trajectories consisting of only 2 to 3 check-ins.}
\end{figure}

\begin{figure}[tbp]
    \centering
    \includegraphics[width=0.39\textwidth]{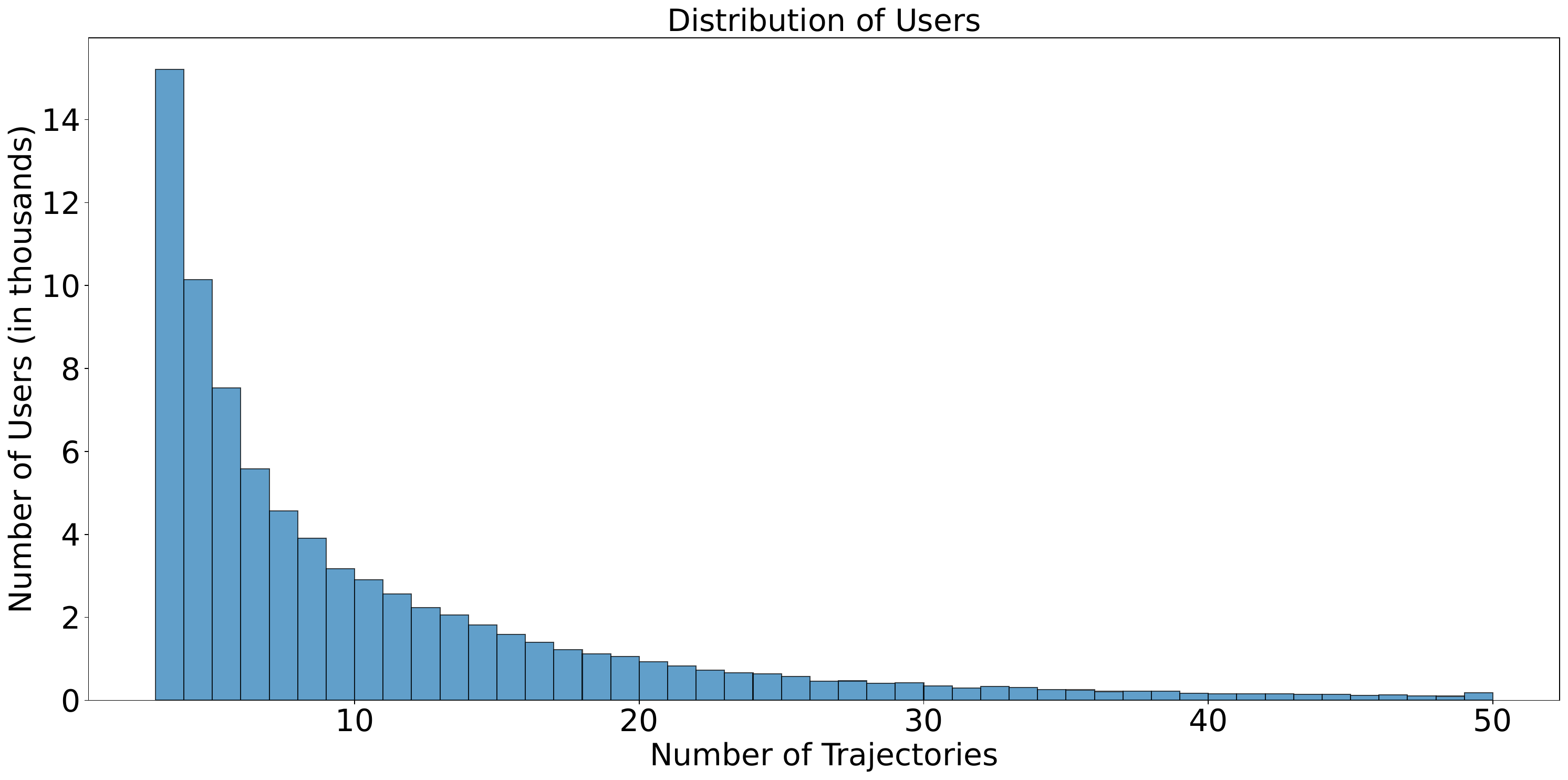}
    \caption{\textbf{Distribution of user activity} based on the number of trajectories per user, indicating a cold-start-heavy dataset.}
    \label{fig:user-activity-distribution}
    \Description{Distribution of user activity based on the number of trajectories per user, indicating a cold-start-heavy dataset.}
\end{figure}

\subsection{Description and Addressing Literature Gaps}

Massive-STEPS is a city-level POI check-in dataset consisting of real-world, user check-in trajectories from 15 cities: Bandung, Beijing, Istanbul, Jakarta, Kuwait City, Melbourne, Moscow, New York, Palembang, Petaling Jaya, São Paulo, Shanghai, Sydney, Tangerang, and Tokyo. It features anonymized POI check-ins enriched with geographical metadata to support spatiotemporal and sequential modeling tasks. City-level statistics, along with comparisons to existing datasets, are presented in Table~\ref{tab:semantic-stats}.

The candidate POI space is substantially larger than that of existing datasets. As shown in Table~\ref{tab:semantic-stats}, datasets like FSQ-NYC and FSQ-TKY \cite{yang2014modeling} contain fewer than 10,000 candidate POI locations. In contrast, cities in Massive-STEPS cover significantly more POIs: New York has over 49,000 POIs, while Jakarta exceeds 76,000. Istanbul, one of the largest subsets, features a large user base of 23,700, offering a broad range of user behaviors. Although some subsets are smaller than their FSQ counterparts (e.g., Tokyo), we attribute this to the strict filtering procedures applied by STD to remove erroneous entries, as explained in Section \ref{sec:existing-datasets}. Analysis of the dataset reveals a long-tailed distribution in both trajectory length (Fig.~\ref{fig:trail-length-distribution}) and user activity (Fig.~\ref{fig:user-activity-distribution}). Most users exhibit cold-start behavior, contributing only a few short trajectories per outing. This prevalence of short and sparse trajectories underscores the necessity for models capable of learning and generalizing effectively from limited sequential data.

Massive-STEPS spans two non-consecutive periods, 2012--2013 and 2017--2018, for 24 months of coverage in total, which enables longitudinal analysis of urban mobility and POI visitation patterns. This temporal span allows us to evaluate how POI recommendation models perform across different time periods (see Section~\ref{sec:longitudinal-experiments}). POIs are inherently dynamic, with substantial closure rates observed in major cities such as New York, Melbourne, and Sydney (see Table~\ref{tab:poi-closures}), indicating that the urban POI landscape is far from static and underscoring the importance of multi-period datasets like ours. Furthermore, we observe clear distributional shifts in the most visited POI categories between the two periods, as illustrated in Fig.~\ref{fig:top-poi-categories-small}, indicating that visitation behaviors can change substantially even within the same city.

Notably, Massive-STEPS includes low-resource and previously underexplored cities in human mobility studies, such as Petaling Jaya and Kuwait City, both of which are among the cities with the highest number of check-ins from STD. This broader coverage opens new research opportunities for studying location-based behaviors across diverse cultural and geographic contexts. Furthermore, since Massive-STEPS is based on STD, it benefits from the carefully filtered, high-quality check-ins and a longer timespan. These characteristics make Massive-STEPS a more relevant and reliable resource for modeling human mobility patterns.

Massive-STEPS is designed to be easily extended to other geographical regions. Since the data processing code is open-source and fully reproducible, adding a new city only requires its geographic boundaries from OpenStreetMap. To ensure full transparency and facilitate further research, all benchmarking scripts and evaluation code used in our experiments are publicly released.

\subsection{Positioning Massive-STEPS}
\label{sec:positioning}

Newer open-source trajectory datasets have been developed with differing objectives and trade-offs. YJMob100K (Japan)~\cite{yabe2024yjmob100k} adopts strong privacy-preserving measures by transforming trajectories into spatial grid cells and discretized time bins, rather than releasing exact GPS coordinates and timestamps. While this design enables mobility research under stricter privacy constraints, it limits analyses that require fine-grained spatial and temporal information. Another direction is the development of synthetic mobility datasets. NUMOSIM (Los Angeles, 2024)~\cite{stanford2024numosim}, WorldMove (2013, 2016-2017, 2022)~\cite{yuan2026worldmove}, and OpenPFLOW (Japan)~\cite{kashiyama2017open} generate synthetic mobility traces through simulation- or model-based processes, enabling controlled experimentation at scales that are often difficult to obtain from real-world data (e.g. anomalous behaviors, urban disruptions). Within this broader mobility dataset landscape, Massive-STEPS combines several characteristics in a single benchmark: (1) large-scale trajectory coverage, (2) real-world mobility traces from LBS check-ins, (3) fine-grained coordinates and timestamps, (4) reproducible construction from publicly available sources, and (5) standardized benchmark splits and evaluation protocols.

\section{Benchmark Tasks}

\subsection{POI Recommendation}

This benchmark focuses on POI recommendation, where the goal is to predict a user's next visit based on their previous check-ins. The input is a trajectory of visited places, and the model is expected to suggest a set of $K$ POIs the user might visit next. It is a supervised task, trained on all available historical trajectories to learn personalized movement patterns. Appendix \ref{sec:suppl-poi-rec} provides details on hyperparameters, experimental setups, and full evaluation results.

\begin{table*}[tbp]
    \centering
    \tiny
    \caption{\textbf{Benchmark results on POI recommendation task}. Full results, including other metrics, are available in Appendix \ref{sec:suppl-poi-rec}. \textbf{Bold} indicates the best performance for each city, while {\ul underline} indicates the second-best.}
    \label{tab:supervised-results}
    \setlength{\tabcolsep}{1.8pt}
    \begin{tabular}{lcc|cc|cc|cc|cc|cc|cc|cc|cc|cc|cc|cc|cc|cc|cc}
        \toprule
        \multirow{2}{*}{\textbf{Model}} & \multicolumn{2}{c}{\textbf{Bandung}} & \multicolumn{2}{c}{\textbf{Beijing}} & \multicolumn{2}{c}{\textbf{Istanbul}} & \multicolumn{2}{c}{\textbf{Jakarta}} & \multicolumn{2}{c}{\textbf{Kuwait City}} & \multicolumn{2}{c}{\textbf{Melbourne}} & \multicolumn{2}{c}{\textbf{Moscow}} & \multicolumn{2}{c}{\textbf{New York}} & \multicolumn{2}{c}{\textbf{Palembang}} & \multicolumn{2}{c}{\textbf{Petaling Jaya}} & \multicolumn{2}{c}{\textbf{São Paulo}} & \multicolumn{2}{c}{\textbf{Shanghai}} & \multicolumn{2}{c}{\textbf{Sydney}} & \multicolumn{2}{c}{\textbf{Tangerang}} & \multicolumn{2}{c}{\textbf{Tokyo}}                                                                                                                                                                                                                                                                \\ \cmidrule{2-31}
                                        & \textbf{A@1}                         & \textbf{N@5}                         & \textbf{A@1}                          & \textbf{N@5}                         & \textbf{A@1}                             & \textbf{N@5}                           & \textbf{A@1}                        & \textbf{N@5}                          & \textbf{A@1}                           & \textbf{N@5}                               & \textbf{A@1}                           & \textbf{N@5}                          & \textbf{A@1}                        & \textbf{N@5}                           & \textbf{A@1}                       & \textbf{N@5}   & \textbf{A@1}   & \textbf{N@5}   & \textbf{A@1}   & \textbf{N@5}   & \textbf{A@1}   & \textbf{N@5}   & \textbf{A@1}   & \textbf{N@5}   & \textbf{A@1}   & \textbf{N@5}   & \textbf{A@1}   & \textbf{N@5}   & \textbf{A@1}   & \textbf{N@5}   \\
        \midrule
        \textbf{FPMC}                   & 0.048                                & 0.083                                & 0.000                                 & 0.009                                & 0.026                                    & 0.050                                  & 0.029                               & 0.058                                 & 0.021                                  & 0.054                                      & 0.062                                  & 0.107                                 & 0.059                               & 0.094                                  & 0.032                              & 0.061          & 0.102          & 0.136          & 0.026          & 0.057          & 0.030          & 0.055          & 0.084          & 0.120          & 0.075          & 0.131          & 0.104          & 0.166          & 0.176          & 0.239          \\
        \midrule
        \textbf{RNN}                    & 0.062                                & 0.099                                & 0.085                                 & 0.134                                & 0.077                                    & 0.130                                  & 0.049                               & 0.083                                 & 0.087                                  & 0.146                                      & 0.059                                  & 0.083                                 & 0.075                               & 0.122                                  & 0.061                              & 0.092          & 0.049          & 0.085          & 0.064          & 0.107          & 0.097          & 0.147          & 0.055          & 0.090          & 0.080          & 0.125          & 0.087          & 0.135          & 0.133          & 0.197          \\
        \textbf{LSTPM}                  & 0.110                                & 0.179                                & 0.127                                 & 0.169                                & 0.142                                    & 0.217                                  & 0.099                               & 0.157                                 & {\ul 0.180}                            & {\ul 0.275}                                & 0.091                                  & 0.150                                 & 0.151                               & 0.229                                  & 0.099                              & 0.155          & 0.114          & 0.175          & 0.099          & 0.163          & 0.158          & 0.243          & 0.099          & 0.149          & 0.141          & 0.206          & 0.154          & 0.237          & {\ul 0.225}    & {\ul 0.315}    \\
        \textbf{DeepMove}               & 0.107                                & 0.172                                & 0.106                                 & 0.190                                & {\ul 0.150}                              & {\ul 0.228}                            & 0.103                               & 0.160                                 & 0.179                                  & 0.274                                      & 0.083                                  & 0.134                                 & 0.143                               & 0.217                                  & 0.097                              & 0.149          & 0.084          & 0.139          & 0.112          & 0.175          & 0.160          & 0.240          & 0.085          & 0.128          & 0.129          & 0.188          & 0.145          & 0.219          & 0.201          & 0.288          \\
        \textbf{Flashback}              & 0.103                                & 0.155                                & 0.278                                 & 0.359                                & 0.137                                    & 0.200                                  & 0.113                               & 0.166                                 & 0.150                                  & 0.227                                      & 0.079                                  & 0.115                                 & 0.121                               & 0.180                                  & 0.086                              & 0.123          & 0.135          & 0.198          & 0.072          & 0.119          & 0.159          & 0.222          & 0.096          & 0.132          & 0.127          & 0.174          & 0.185          & 0.262          & 0.170          & 0.237          \\
        \textbf{Graph-Flashback}        & 0.091                                & 0.146                                & 0.304                                 & 0.373                                & 0.137                                    & 0.202                                  & 0.072                               & 0.121                                 & 0.115                                  & 0.181                                      & 0.093                                  & 0.143                                 & 0.125                               & 0.188                                  & 0.059                              & 0.093          & 0.148          & 0.208          & 0.072          & 0.119          & 0.096          & 0.156          & 0.111          & 0.151          & 0.135          & 0.188          & 0.200          & 0.292          & 0.180          & 0.259          \\
        \textbf{LoTNext}                & 0.076                                & 0.125                                & 0.243                                 & 0.320                                & 0.105                                    & 0.168                                  & 0.044                               & 0.078                                 & 0.132                                  & 0.209                                      & 0.085                                  & 0.133                                 & 0.089                               & 0.145                                  & 0.061                              & 0.099          & 0.140          & 0.208          & 0.055          & 0.097          & 0.105          & 0.166          & 0.093          & 0.139          & 0.123          & 0.176          & 0.126          & 0.190          & 0.191          & 0.258          \\
        \midrule
        \textbf{GETNext}                & {\ul 0.179}                          & {\ul 0.247}                          & {\ul 0.433}                           & {\ul 0.486}                          & 0.146                                    & 0.210                                  & {\ul 0.155}                         & {\ul 0.209}                           & 0.175                                  & 0.251                                      & {\ul 0.100}                            & {\ul 0.179}                           & {\ul 0.175}                         & {\ul 0.260}                            & {\ul 0.134}                        & {\ul 0.202}    & {\ul 0.158}    & {\ul 0.239}    & {\ul 0.139}    & {\ul 0.200}    & {\ul 0.202}    & {\ul 0.286}    & {\ul 0.115}    & {\ul 0.177}    & {\ul 0.181}    & {\ul 0.266}    & {\ul 0.224}    & {\ul 0.302}    & 0.180          & 0.275          \\
        \textbf{STHGCN}                 & \textbf{0.219}                       & \textbf{0.302}                       & \textbf{0.453}                        & \textbf{0.552}                       & \textbf{0.241}                           & \textbf{0.318}                         & \textbf{0.197}                      & \textbf{0.270}                        & \textbf{0.225}                         & \textbf{0.314}                             & \textbf{0.168}                         & \textbf{0.247}                        & \textbf{0.223}                      & \textbf{0.308}                         & \textbf{0.146}                     & \textbf{0.207} & \textbf{0.246} & \textbf{0.341} & \textbf{0.174} & \textbf{0.241} & \textbf{0.250} & \textbf{0.344} & \textbf{0.193} & \textbf{0.264} & \textbf{0.227} & \textbf{0.307} & \textbf{0.293} & \textbf{0.400} & \textbf{0.250} & \textbf{0.350} \\
        \midrule
        \textbf{GNPR-SID}               & 0.126                                & 0.136                                & 0.139                                 & 0.143                                & 0.115                                    & 0.127                                  & 0.102                               & 0.107                                 & 0.162                                  & 0.178                                      & 0.049                                  & 0.054                                 & 0.116                               & 0.128                                  & 0.074                              & 0.080          & 0.094          & 0.103          & 0.090          & 0.098          & 0.165          & 0.176          & 0.060          & 0.067          & 0.105          & 0.112          & 0.165          & 0.177          & 0.114          & 0.126          \\
        \textbf{UniMove}                & 0.007                                & 0.033                                & 0.036                                 & 0.128                                & 0.015                                    & 0.038                                  & 0.004                               & 0.020                                 & 0.023                                  & 0.073                                      & 0.008                                  & 0.037                                 & 0.009                               & 0.030                                  & 0.004                              & 0.016          & 0.009          & 0.035          & 0.008          & 0.034          & 0.002          & 0.009          & 0.000          & 0.029          & 0.015          & 0.059          & 0.001          & 0.029          & 0.032          & 0.072          \\
        \bottomrule
    \end{tabular}
\end{table*}

\subsubsection{Problem Formulation}

We adopt the conventional problem formulation used in prior POI recommendation studies \cite{zhang2025survey, 10.1145/3477495.3531983, 10.1145/3539618.3591770}, which defines the task as learning user preferences and routines from historical check-ins to recommend future POIs. Let $\mathcal{U} = \{u_1, u_2, \dots, u_M\}$ denote the set of users, $\mathcal{P} = \{p_1, p_2, \dots, p_N\}$ the set of Points of Interest (POIs), and $\mathcal{T} = \{t_1, t_2, \dots, t_K\}$ the set of timestamps, where $M, N, K \in \mathbb{N}$.
A POI $p \in \mathcal{P}$ is represented as:
\[
    p = \langle \phi, \lambda, \kappa, \alpha, \beta, \gamma \rangle,
\]
where $\phi$ and $\lambda$ are the latitude and longitude, $\kappa$ is the POI category (e.g., \textit{restaurant}, \textit{park}), $\alpha$ is the unique POI identifier, $\beta$ is the textual address, and $\gamma$ is the POI name.
A check-in is a tuple $c = \langle u, p, t \rangle \in \mathcal{U} \times \mathcal{P} \times \mathcal{T}$, indicating that user $u$ visited POI $p$ at timestamp $t$.
A trajectory for user $u$ is defined as a temporally ordered sequence of check-ins within a fixed time interval $\delta \tau = 8$ hours. Each trajectory $T^i_u(t)$ up to timestamp $t$ is defined as:
\[
    T^i_u(t) = \{(p_1, t_1), (p_2, t_2), \dots, (p_k, t_k)\}
\]
such that $t_1 < t_2 < \dots < t_k = t$ and $t_k - t_{k-1} \leq \delta \tau$. Given a set of historical trajectories:
\[
    \mathcal{T}_u = \{T^1_u, T^2_u, \dots, T^L_u\}
\]
for user $u$, where $L$ is the number of such \textbf{historical} trajectories, the goal is to recommend the POIs that $u$ is most likely to visit next after the current \textbf{contextual} trajectory $T'_u(t)$.


Given a current contextual trajectory $T'_u(t)$ of user $u$ up to time $t$, along with their historical trajectories $\mathcal{T}_u$, the task of next POI recommendation is to rank all candidate POIs $p_i \in \mathcal{P}$ according to the model's predicted probability that user $u$ will visit each POI next.
Formally, the model learns a ranking function:
\[
    f: (T'_u(t), \mathcal{T}_u) \rightarrow \{ \hat{y}_i \}_{i=1}^{|\mathcal{P}|}
\]
where $\hat{y}_i$ denotes the predicted likelihood that user $u$ will visit POI $p_i$ next. Based on these scores, a ranked list of POIs is returned as recommendations.
This formulation enables POI recommendation, where the goal is to suggest a set of likely POIs that a user may visit next based on their historical check-ins. Our evaluation metrics, Acc@k and NDCG@k, assess whether the ground-truth POI appears among the top-$k$ ranked candidates, reflecting the quality of the recommended set. In particular, Acc@1 captures the stricter task of \textit{immediate} next POI prediction, measuring whether the top-ranked POI matches the user's actual next visit.

\begin{table*}[tbp]
    \centering
    \tiny
    \caption{\textbf{Benchmark results on zero-shot POI recommendation task}. Full results, including other metrics, are available in Appendix \ref{sec:suppl-poi-rec-zs}. \textbf{Bold} indicates the best performance for each city, while {\ul underline} indicates the second-best.}
    \label{tab:zero-shot-results}
    \setlength{\tabcolsep}{1.3pt}
    \begin{tabular}{llcc|cc|cc|cc|cc|cc|cc|cc|cc|cc|cc|cc|cc|cc|cc}
        \toprule
        \multicolumn{2}{l}{\multirow{2}{*}{\textbf{Method}}} & \multicolumn{2}{c}{\textbf{Bandung}} & \multicolumn{2}{c}{\textbf{Beijing}} & \multicolumn{2}{c}{\textbf{Istanbul}} & \multicolumn{2}{c}{\textbf{Jakarta}} & \multicolumn{2}{c}{\textbf{Kuwait City}} & \multicolumn{2}{c}{\textbf{Melbourne}} & \multicolumn{2}{c}{\textbf{Moscow}} & \multicolumn{2}{c}{\textbf{New York}} & \multicolumn{2}{c}{\textbf{Palembang}} & \multicolumn{2}{c}{\textbf{Petaling Jaya}} & \multicolumn{2}{c}{\textbf{São Paulo}} & \multicolumn{2}{c}{\textbf{Shanghai}} & \multicolumn{2}{c}{\textbf{Sydney}} & \multicolumn{2}{c}{\textbf{Tangerang}} & \multicolumn{2}{c}{\textbf{Tokyo}}                                                                                                                                                                                                                                                                                 \\ \cmidrule{3-32}
        \multicolumn{2}{l}{}                                 & \textbf{A@1}                         & \textbf{N@5}                         & \textbf{A@1}                          & \textbf{N@5}                         & \textbf{A@1}                             & \textbf{N@5}                           & \textbf{A@1}                        & \textbf{N@5}                          & \textbf{A@1}                           & \textbf{N@5}                               & \textbf{A@1}                           & \textbf{N@5}                          & \textbf{A@1}                        & \textbf{N@5}                           & \textbf{A@1}                       & \textbf{N@5}   & \textbf{A@1}   & \textbf{N@5}   & \textbf{A@1}   & \textbf{N@5}   & \textbf{A@1}   & \textbf{N@5}   & \textbf{A@1}   & \textbf{N@5}   & \textbf{A@1}   & \textbf{N@5}   & \textbf{A@1}   & \textbf{N@5}   & \textbf{A@1}   & \textbf{N@5}                    \\
        \midrule
        \multirow{5}{*}{\textbf{LLM-Mob}}                    & \textbf{Gemini 2 Flash}              & 0.105                                & 0.139                                 & 0.115                                & 0.226                                    & 0.080                                  & 0.160                               & 0.100                                 & 0.174                                  & 0.095                                      & 0.185                                  & 0.060                                 & 0.108                               & 0.130                                  & 0.187                              & 0.095          & 0.136          & 0.135          & 0.208          & 0.090          & 0.160          & 0.130          & 0.223          & 0.055          & 0.111          & 0.060          & 0.112          & 0.155          & 0.225          & 0.140          & 0.238          \\
                                                             & \textbf{Claude Haiku 4.5}            & 0.115                                & 0.150                                 & 0.115                                & 0.268                                    & 0.085                                  & 0.182                               & 0.135                                 & 0.201                                  & 0.110                                      & 0.190                                  & 0.070                                 & 0.113                               & 0.130                                  & 0.203                              & 0.075          & 0.128          & 0.135          & 0.214          & 0.090          & 0.156          & 0.170          & 0.251          & 0.070          & 0.122          & 0.075          & 0.119          & 0.155          & 0.244          & 0.180          & 0.298          \\
                                                             & \textbf{Qwen 2.5 7B}                 & 0.060                                & 0.111                                 & 0.058                                & 0.218                                    & 0.035                                  & 0.148                               & 0.105                                 & 0.179                                  & 0.080                                      & 0.155                                  & 0.030                                 & 0.083                               & 0.090                                  & 0.185                              & 0.070          & 0.131          & 0.075          & 0.143          & 0.030          & 0.116          & 0.090          & 0.188          & 0.040          & 0.108          & 0.035          & 0.091          & 0.095          & 0.196          & 0.110          & 0.243          \\
                                                             & \textbf{Llama 3.1 8B}                & 0.010                                & 0.055                                 & 0.000                                & 0.000                                    & 0.020                                  & 0.065                               & 0.055                                 & 0.104                                  & 0.030                                      & 0.066                                  & 0.010                                 & 0.040                               & 0.030                                  & 0.068                              & 0.025          & 0.061          & 0.005          & 0.025          & 0.010          & 0.050          & 0.030          & 0.098          & 0.005          & 0.013          & 0.020          & 0.053          & 0.020          & 0.073          & 0.005          & 0.025          \\
                                                             & \textbf{Gemma 2 9B}                  & 0.070                                & 0.126                                 & 0.115                                & 0.206                                    & 0.075                                  & 0.146                               & 0.105                                 & 0.178                                  & 0.080                                      & 0.150                                  & 0.055                                 & 0.108                               & 0.100                                  & 0.176                              & 0.070          & 0.124          & 0.095          & 0.171          & 0.055          & 0.122          & 0.085          & 0.162          & 0.050          & 0.104          & 0.030          & 0.086          & 0.145          & 0.209          & 0.145          & 0.255          \\
        \midrule
        \multirow{5}{*}{\textbf{LLM-ZS}}                     & \textbf{Gemini 2 Flash}              & 0.095                                & 0.147                                 & 0.058                                & 0.246                                    & 0.090                                  & 0.166                               & 0.110                                 & 0.188                                  & 0.080                                      & 0.167                                  & 0.065                                 & 0.115                               & 0.125                                  & 0.217                              & 0.080          & 0.129          & 0.130          & 0.196          & 0.110          & 0.164          & 0.150          & 0.235          & 0.065          & 0.113          & 0.060          & 0.111          & 0.145          & 0.234          & 0.160          & 0.278          \\
                                                             & \textbf{Claude Haiku 4.5}            & 0.115                                & 0.156                                 & 0.115                                & {\ul 0.273}                              & 0.100                                  & 0.178                               & 0.145                                 & 0.206                                  & 0.130                                      & 0.207                                  & 0.070                                 & 0.107                               & 0.130                                  & 0.206                              & 0.090          & 0.141          & 0.165          & 0.227          & 0.120          & 0.174          & 0.170          & 0.251          & 0.075          & 0.128          & 0.070          & 0.116          & 0.175          & 0.254          & 0.205          & 0.306          \\
                                                             & \textbf{Qwen 2.5 7B}                 & 0.055                                & 0.126                                 & 0.038                                & 0.237                                    & 0.040                                  & 0.141                               & 0.065                                 & 0.161                                  & 0.050                                      & 0.140                                  & 0.040                                 & 0.100                               & 0.080                                  & 0.176                              & 0.050          & 0.116          & 0.050          & 0.135          & 0.045          & 0.111          & 0.095          & 0.198          & 0.045          & 0.103          & 0.045          & 0.109          & 0.100          & 0.215          & 0.120          & 0.257          \\
                                                             & \textbf{Llama 3.1 8B}                & 0.045                                & 0.131                                 & 0.077                                & 0.221                                    & 0.040                                  & 0.137                               & 0.045                                 & 0.126                                  & 0.060                                      & 0.137                                  & 0.040                                 & 0.101                               & 0.080                                  & 0.183                              & 0.055          & 0.111          & 0.070          & 0.154          & 0.030          & 0.123          & 0.030          & 0.159          & 0.060          & 0.116          & 0.040          & 0.110          & 0.080          & 0.173          & 0.110          & 0.269          \\
                                                             & \textbf{Gemma 2 9B}                  & 0.065                                & 0.130                                 & 0.096                                & 0.217                                    & 0.045                                  & 0.141                               & 0.105                                 & 0.180                                  & 0.070                                      & 0.153                                  & 0.050                                 & 0.100                               & 0.080                                  & 0.194                              & 0.075          & 0.129          & 0.060          & 0.150          & 0.065          & 0.126          & 0.075          & 0.192          & 0.050          & 0.112          & 0.045          & 0.103          & 0.100          & 0.227          & 0.110          & 0.263          \\
        \midrule
        \multirow{5}{*}{\textbf{LLM-Move}}                   & \textbf{Gemini 2 Flash}              & \textbf{0.225}                       & \textbf{0.289}                        & 0.096                                & 0.218                                    & \textbf{0.205}                         & {\ul 0.289}                         & \textbf{0.295}                        & \textbf{0.350}                         & \textbf{0.220}                             & \textbf{0.295}                         & \textbf{0.225}                        & \textbf{0.275}                      & {\ul 0.220}                            & {\ul 0.316}                        & \textbf{0.235} & \textbf{0.325} & \textbf{0.260} & \textbf{0.329} & \textbf{0.210} & \textbf{0.273} & \textbf{0.285} & \textbf{0.350} & \textbf{0.170} & \textbf{0.221} & {\ul 0.230}    & \textbf{0.331} & {\ul 0.200}    & {\ul 0.274}    & {\ul 0.250}    & \textbf{0.368} \\
                                                             & \textbf{Claude Haiku 4.5}            & {\ul 0.205}                          & {\ul 0.240}                           & {\ul 0.154}                          & 0.227                                    & {\ul 0.195}                            & \textbf{0.295}                      & {\ul 0.260}                           & {\ul 0.287}                            & {\ul 0.185}                                & {\ul 0.283}                            & {\ul 0.175}                           & {\ul 0.249}                         & \textbf{0.230}                         & \textbf{0.330}                     & {\ul 0.170}    & {\ul 0.241}    & 0.175          & 0.195          & 0.165          & {\ul 0.208}    & {\ul 0.275}    & {\ul 0.325}    & 0.100          & {\ul 0.141}    & \textbf{0.235} & {\ul 0.325}    & \textbf{0.220} & \textbf{0.275} & \textbf{0.255} & {\ul 0.340}    \\
                                                             & \textbf{Qwen 2.5 7B}                 & 0.100                                & 0.128                                 & \textbf{0.192}                       & \textbf{0.280}                           & 0.175                                  & 0.226                               & 0.115                                 & 0.169                                  & 0.160                                      & 0.227                                  & 0.110                                 & 0.165                               & \textbf{0.230}                         & 0.274                              & 0.120          & 0.188          & 0.130          & 0.163          & 0.135          & 0.155          & 0.155          & 0.199          & 0.095          & 0.133          & 0.125          & 0.205          & 0.175          & 0.229          & {\ul 0.250}    & 0.312          \\
                                                             & \textbf{Llama 3.1 8B}                & 0.030                                & 0.033                                 & 0.058                                & 0.100                                    & 0.015                                  & 0.036                               & 0.015                                 & 0.021                                  & 0.010                                      & 0.023                                  & 0.040                                 & 0.123                               & 0.005                                  & 0.031                              & 0.035          & 0.084          & 0.010          & 0.013          & 0.040          & 0.049          & 0.045          & 0.045          & 0.020          & 0.030          & 0.055          & 0.141          & 0.000          & 0.003          & 0.030          & 0.046          \\
                                                             & \textbf{Gemma 2 9B}                  & 0.175                                & 0.213                                 & 0.096                                & 0.229                                    & 0.100                                  & 0.155                               & 0.235                                 & 0.266                                  & 0.120                                      & 0.202                                  & 0.115                                 & 0.199                               & 0.110                                  & 0.185                              & 0.115          & 0.183          & {\ul 0.210}    & {\ul 0.240}    & {\ul 0.175}    & {\ul 0.208}    & 0.195          & 0.252          & {\ul 0.105}    & 0.128          & 0.125          & 0.254          & 0.125          & 0.193          & 0.130          & 0.225          \\
        \bottomrule
    \end{tabular}
\end{table*}

\subsubsection{Experimental Setup}

We adopted the predefined trajectories from the original STD, where check-ins are grouped into sequences based on fixed time intervals (see Section \ref{sec:data-preprocessing}). All input features are numerically encoded, enabling straightforward use across experiments. Models typically use four feature types: (1) social: user ID; (2) spatial: POI ID and geographic coordinates; (3) temporal: check-in timestamp; and (4) categorical: POI category. For POIs lacking exact geographic coordinates due to privacy (see Section \ref{sec:poi-enrichment}), we defaulted to the coordinates of their administrative region as a spatial proxy. We evaluated 11 baseline models:

\begin{itemize}
    \item \textbf{FPMC} \cite{10.1145/1772690.1772773}: A classical baseline that combines first-order Markov chains with matrix factorization to model personalized next-location predictions.
    \item \textbf{RNN} \cite{10.1145/3474717.3483923}, \textbf{LSTPM} \cite{Sun_Qian_Chen_Liang_Nguyen_Yin_2020}, \textbf{DeepMove} \cite{10.1145/3178876.3186058}, \textbf{Flashback} \cite{10.5555/3491440.3491742}, \textbf{Graph-Flashback} \cite{10.1145/3534678.3539383}, \textbf{LoTNext} \cite{10.5555/3737916.3739652}: Recurrent neural networks designed to capture sequential dependencies, with mechanisms to incorporate spatio-temporal context.
    \item \textbf{GETNext} \cite{10.1145/3477495.3531983}, \textbf{STHGCN} \cite{10.1145/3539618.3591770}: Transformer-based graph neural networks to model spatial, and temporal dependencies.
    \item \textbf{GNPR-SID} \cite{10.1145/3711896.3736981}: A large language model fine-tuning approach that formulates POI recommendation as a next semantic token prediction task.
    \item \textbf{UniMove} \cite{han2025unimoveunifiedmodelmulticity}: Transformer-based trajectory foundation model with Mixture of Experts layers.
\end{itemize}
We employed two commonly used metrics in POI recommender systems: Acc@k, which checks if the true POI appears in the top-k predicted results, and NDCG@k, which measures the ranking quality of the suggested results.


\subsubsection{Main Results}

As shown in Table~\ref{tab:supervised-results}, STHGCN achieves the highest average Acc@1 and NDCG@5 across all cities, followed closely by GETNext, demonstrating the effectiveness of GNNs compared to classical recurrent models. The top model attained a mean Acc@1 of 23.4\%, comparable to previous studies on similarly sized datasets \cite{feng-etal-2025-agentmove}. Interestingly, newer variants of Flashback, such as Graph-Flashback and LoTNext, do not always provide a significant improvement, contrary to their original claims. We note that in their respective studies, these methods were evaluated only on two datasets (Gowalla and Foursquare); our results further highlight the importance of benchmarking breadth to ensure that improvements generalize across datasets and diverse urban environments. Similarly, pre-training UniMove from scratch (pre-trained weights are not publicly available), struggled to surpass older baselines. We attribute this to the high number of cold-start trajectories (see Fig.~\ref{fig:trail-length-distribution}), as the method fails to generalize to the extremely short input sequences typical of real-world data. In these cases, next-token prediction loss struggles to provide meaningful training signal.

\subsubsection{Correlating Urban Features with Model Scores}
We also examined the impact of urban features on POI recommendation accuracy by computing Spearman correlations between city features and model performance.
Prior studies \cite{10.1145/3477495.3531983,10.1145/3539618.3591770,10.1145/3626772.3657840} have suggested several features as potential explanatory variables, including:

\begin{itemize}
    \item Number of unique check-ins,
    \item Number of unique trajectories,
    \item Number of unique POI categories,
    \item Geographical area (larger areas are assumed to be harder to model), and
    \item POI or spatial sparsity (i.e., unique POIs per unit area).
\end{itemize}
We also propose several additional features for consideration:
\begin{itemize}
    \item Number of unique POIs,
    \item Check-in density (unique check-ins per area),
    \item Trajectory density (unique trajectories per area), and
    \item Category entropy, our proposed metric capturing categorical diversity.
\end{itemize}

\textbf{Category entropy}, based on Shannon entropy, measures how evenly POI categories are distributed in a city. A higher entropy suggests that check-ins are spread across a wide variety of categories, while a lower entropy indicates a concentration in fewer venue types. It is defined as $H = -\sum_{i=1}^{N} p_i \log(p_i)$, where $N$ is the total number of POI categories and $p_i = c_i / \sum_{j=1}^{N} c_j$ is the fraction of all venues belonging to category $i$, with $c_i$ the count of venues in that category.

\begin{figure}[bp]
    \centering
    \includegraphics[width=0.44\textwidth]{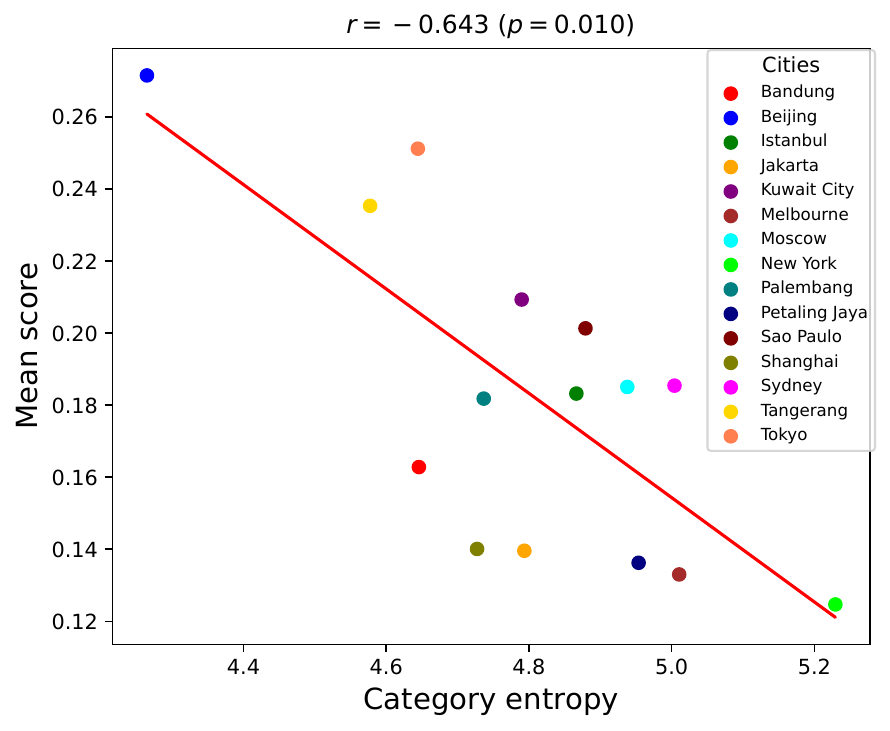}
    \caption{Spearman correlation between category entropy and mean score of POI recommendation models.}
    \label{fig:category-entropy}
    \Description{Spearman correlation between category entropy and mean score of POI recommendation models.}
\end{figure}

Moreover, previous studies have primarily focused on only three datasets: FSQ-NYC, FSQ-TKY, and Gowalla-CA. In contrast, Massive-STEPS provides broader coverage across 15 cities, enabling a more comprehensive and robust analysis. To examine the relationship between urban features and model performance, we averaged the three evaluation metrics across nine supervised baselines for each city and computed the Spearman correlation with each candidate feature. 
Among all features, \textbf{category entropy} shows the strongest correlation with model performance, with a Spearman correlation of $r = -0.643$ ($p < 0.05$) (see Fig.~\ref{fig:category-entropy}). This suggests that cities with more evenly distributed POI categories \textit{tend} to yield lower prediction accuracy. Intuitively, when no single category dominates (a city has roughly equal proportions of restaurants, cafes, homes, and other POIs), it becomes more difficult for models to predict a user's next destination. In these cases, user behavior is more varied and less predictable. On the other hand, cities with more skewed category distributions (e.g., mostly food places or mostly residential areas) tend to have more consistent patterns of movement, making them easier for models to learn and predict. Interestingly, our finding contradicts the hypothesis of \cite{10.1145/3626772.3657840}, which suggests that FSQ-NYC is easier to model than Gowalla-CA due to the former's vast number of POI categories that provide richer contextual signals for the model. Our results indicate that it is not the number of categories that matters, but rather how these categories are distributed.


\begin{table*}[tbp]
    \centering
    \footnotesize
    \caption{\textbf{Zero-shot POI recommendation results using LLM-Move across two time periods.} The metric reported is Acc@1. \textbf{Bold} indicates the best performance for each city, while {\ul underline} indicates the second-best. N/A indicates that no samples were available for that city in the corresponding time period.}
    \label{tab:longitudinal-zero-shot-results}
    \setlength{\tabcolsep}{1.3pt}
    \begin{tabular}{clccccccccccccccc}
        \toprule
        \multicolumn{1}{l}{\textbf{Time Period}} & \textbf{Model}            & \textbf{Bandung} & \textbf{Beijing} & \textbf{Istanbul} & \textbf{Jakarta} & \textbf{KC}    & \textbf{Melbourne} & \textbf{Moscow} & \textbf{NY}    & \textbf{Palembang} & \textbf{PJ}    & \textbf{SP}    & \textbf{Shanghai} & \textbf{Sydney} & \textbf{Tangerang} & \textbf{Tokyo} \\
        \midrule
        \multirow{5}{*}{\textbf{2012-2013}}      & \textbf{Gemini 2 Flash}   & \textbf{0.227}   & 0.102            & \textbf{0.212}    & \textbf{0.295}   & \textbf{0.423} & \textbf{0.226}     & {\ul 0.218}     & \textbf{0.240} & \textbf{0.256}     & \textbf{0.199} & \textbf{0.298} & \textbf{0.192}    & \textbf{0.256}  & {\ul 0.197}        & N/A            \\
                                                 & \textbf{Claude Haiku 4.5} & {\ul 0.206}      & {\ul 0.163}      & {\ul 0.192}       & {\ul 0.259}      & 0.231          & {\ul 0.168}        & \textbf{0.234}  & {\ul 0.182}    & 0.174              & {\ul 0.184}    & {\ul 0.286}    & 0.103             & {\ul 0.244}     & \textbf{0.217}     & N/A            \\
                                                 & \textbf{Qwen 2.5 7B}      & 0.098            & \textbf{0.204}   & {\ul 0.192}       & 0.114            & 0.269          & 0.116              & \textbf{0.234}  & 0.130          & 0.128              & 0.142          & 0.173          & {\ul 0.109}       & 0.122           & 0.172              & N/A            \\
                                                 & \textbf{Llama 3.1 8B}     & 0.031            & 0.041            & 0.007             & 0.010            & 0.000          & 0.039              & 0.005           & 0.032          & 0.010              & 0.014          & 0.048          & 0.006             & 0.064           & 0.000              & N/A            \\
                                                 & \textbf{Gemma 2 9B}       & 0.180            & {\ul 0.102}      & 0.116             & 0.228            & {\ul 0.308}    & 0.097              & 0.112           & 0.130          & {\ul 0.215}        & \textbf{0.199} & 0.202          & {\ul 0.109}       & 0.122           & 0.126              & N/A            \\
        \midrule
        \multirow{5}{*}{\textbf{2017-2018}}      & \textbf{Gemini 2 Flash}   & \textbf{0.167}   & 0.000            & {\ul 0.185}       & {\ul 0.286}      & \textbf{0.190} & \textbf{0.222}     & \textbf{0.333}  & \textbf{0.217} & \textbf{0.400}     & \textbf{0.237} & \textbf{0.219} & \textbf{0.091}    & {\ul 0.136}     & \textbf{0.500}     & {\ul 0.250}    \\
                                                 & \textbf{Claude Haiku 4.5} & \textbf{0.167}   & 0.000            & \textbf{0.204}    & {\ul 0.286}      & {\ul 0.178}    & {\ul 0.200}        & 0.000           & {\ul 0.130}    & {\ul 0.200}        & {\ul 0.119}    & \textbf{0.219} & \textbf{0.091}    & \textbf{0.205}  & \textbf{0.500}     & \textbf{0.255} \\
                                                 & \textbf{Qwen 2.5 7B}      & \textbf{0.167}   & 0.000            & 0.130             & 0.143            & 0.144          & 0.089              & 0.000           & 0.087          & {\ul 0.200}        & {\ul 0.119}    & 0.063          & 0.045             & {\ul 0.136}     & \textbf{0.500}     & {\ul 0.250}    \\
                                                 & \textbf{Llama 3.1 8B}     & 0.000            & \textbf{0.333}   & 0.037             & 0.143            & 0.011          & 0.044              & 0.000           & 0.043          & 0.000              & 0.102          & 0.031          & {\ul 0.068}       & {\ul 0.023}     & 0.000              & 0.030          \\
                                                 & \textbf{Gemma 2 9B}       & 0.000            & 0.000            & 0.056             & \textbf{0.429}   & 0.092          & 0.178              & 0.000           & 0.065          & 0.000              & {\ul 0.119}    & {\ul 0.156}    & \textbf{0.091}    & {\ul 0.136}     & 0.000              & 0.130          \\
        \bottomrule
    \end{tabular}
\end{table*}

\subsection{Zero-shot POI Recommendation}

This benchmark focuses on zero-shot POI recommendation via LLMs, where the goal is to predict a user's next visit based on their previous check-ins (similar to its supervised counterpart) without additional model fine-tuning. The input is a user trajectory transformed into a textual prompt, and the model ranks a set of $K$ candidate POIs to identify the next likely destination. Appendix \ref{sec:suppl-poi-rec-zs} provides details on experimental setups, and full evaluation results.

\subsubsection{Problem Formulation}

The zero-shot POI recommendation task follows the same problem formulation as its supervised counterpart. The key difference is that in this setting, the model parameters remain frozen and the models are pre-trained, rather than trained from randomly initialized weights.

\subsubsection{Experimental Setup}

For zero-shot recommendation, trajectories are converted into textual prompts \cite{10.1145/3557915.3561026,10.1109/TKDE.2023.3342137}. We adapted the prompt templates from \cite{feng-etal-2025-agentmove}, which implemented the three LLM methods evaluated in this study:

\begin{itemize}
    \item \textbf{LLM-Mob} \cite{wang2023would}: One of the earliest methods to use LLMs for next POI prediction, prompting LLMs with both historical and current (contextual) trajectories.
    \item \textbf{LLM-ZS} \cite{beneduce2024large}: A simplified version of LLM-Mob that retains the use of historical and contextual trajectories but simplifies its prompt design.
    \item \textbf{LLM-Move} \cite{feng2024move}: Extends previous prompting methods by introducing a RAG-like approach, retrieving random POIs as candidates, and ranking them by geographic distance to the user's most recent visit.
\end{itemize}


Since LLMs can leverage contextual information, features do not need numerical encoding; we used each check-in's timestamp, POI category name, and POI ID. For a robust evaluation, we tested each method on four LLMs: two closed-source APIs: Gemini 2.0 Flash \cite{geminiteam2024geminifamilyhighlycapable} and Claude Haiku 4.5, and three open-source instruction-tuned models: Qwen 2.5 7B \cite{qwen2.5}, Llama 3.1 8B \cite{grattafiori2024llama3herdmodels}, and Gemma 2 9B \cite{gemmateam2024gemma2improvingopen}. We used the same metrics as in the supervised setting.

\begin{figure}[bp]
    \centering
    \includegraphics[width=0.47\textwidth]{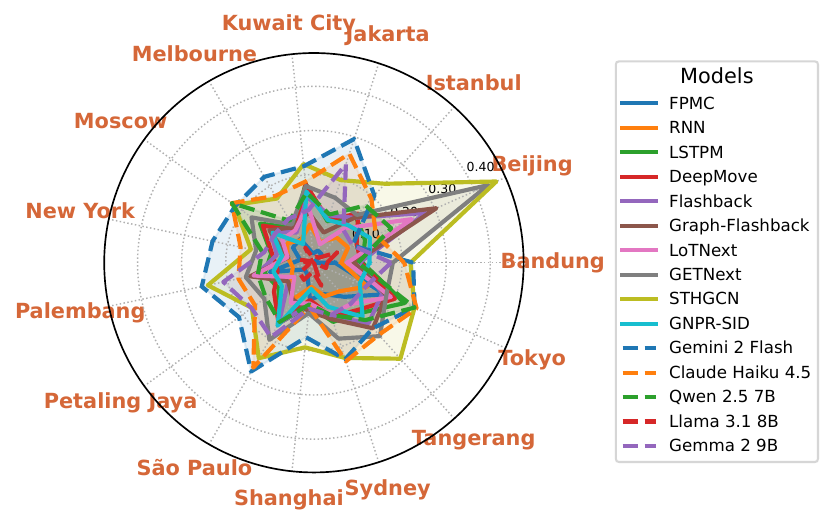}
    \caption{Acc@1 of supervised (solid lines) and LLM-Move (dashed lines) models across 15 cities.}
    \label{fig:model-radar}
    \Description{Acc@1 of supervised (solid lines) and LLM-Move (dashed lines) models across 15 cities.}
\end{figure}

\subsubsection{Main Results}

As shown in Table \ref{tab:zero-shot-results}, LLM-Move \cite{feng2024move} outperformed the other two methods due to its prompt design which provides candidate POIs rather than relying solely on historical or contextual trajectories, unlike LLM-Mob and LLM-ZS. Importantly, LLM-Move ranks candidate locations based on spatial distance from the user's latest check-in, echoing the approach of TimeGeo~\cite{doi:10.1073/pnas.1524261113}. Across LLMs, Gemini 2.0 Flash achieved the highest accuracy across all prompting strategies, followed closely by Claude Haiku 4.5, with Qwen 2.5 7B and Gemma 2 9B as strong open-source alternatives. Notably, as shown in Fig.~\ref{fig:model-radar}, these zero-shot methods matched or exceeded supervised baselines in several cities (e.g., Jakarta, Kuwait City, Moscow), demonstrating their effectiveness without fine-tuning. 

\subsubsection{Longitudinal Experiments and Results}
\label{sec:longitudinal-experiments}

\begin{table*}[tbp]
    \centering
    \footnotesize
    \caption{\textbf{Benchmark results on spatiotemporal classification task}. The metric reported is Accuracy. \textbf{Bold} indicates the best performance for each city, while {\ul underline} indicates the second-best.}
    \label{tab:st-day-results}
    \setlength{\tabcolsep}{1.5pt}
    \begin{tabular}{lccccccccccccccc}
        \toprule
        \textbf{Model}                                 & \textbf{Bandung} & \textbf{Beijing} & \textbf{Istanbul} & \textbf{Jakarta} & \textbf{KC}    & \textbf{Melbourne} & \textbf{Moscow} & \textbf{NY}    & \textbf{Palembang} & \textbf{PJ}    & \textbf{SP}    & \textbf{Shanghai} & \textbf{Sydney} & \textbf{Tangerang} & \textbf{Tokyo} \\
        \midrule
        \textbf{Logistic Regression} (\textit{BoW})    & {\ul 0.645}      & \textbf{0.750}   & 0.645             & \textbf{0.665}   & {\ul 0.785}    & 0.700              & 0.715           & 0.550          & 0.630              & 0.565          & 0.710          & 0.585             & \textbf{0.680}  & 0.625              & \textbf{0.595} \\
        \textbf{Logistic Regression} (\textit{TF-IDF}) & 0.620            & \textbf{0.750}   & 0.650             & {\ul 0.660}      & {\ul 0.785}    & {\ul 0.705}        & 0.720           & 0.555          & 0.665              & 0.560          & {\ul 0.735}    & \textbf{0.600}    & {\ul 0.650}     & 0.645              & 0.570          \\
        \textbf{Random Forest} (\textit{BoW})          & 0.615            & {\ul 0.692}      & 0.665             & 0.635            & {\ul 0.785}    & \textbf{0.710}     & {\ul 0.740}     & 0.570          & \textbf{0.680}     & 0.580          & 0.715          & \textbf{0.600}    & 0.645           & {\ul 0.675}        & 0.550          \\
        \textbf{Random Forest} (\textit{TF-IDF})       & 0.615            & 0.673            & 0.665             & 0.635            & {\ul 0.785}    & {\ul 0.705}        & {\ul 0.740}     & 0.565          & \textbf{0.680}     & {\ul 0.590}    & 0.715          & \textbf{0.600}    & {\ul 0.650}     & \textbf{0.685}     & 0.510          \\
        \textbf{XGBoost} (\textit{BoW})                & 0.620            & 0.577            & 0.600             & 0.635            & 0.745          & 0.695              & 0.670           & 0.570          & 0.625              & 0.580          & 0.645          & \textbf{0.600}    & 0.600           & 0.615              & 0.565          \\
        \textbf{XGBoost} (\textit{TF-IDF})             & \textbf{0.660}   & 0.615            & 0.610             & 0.640            & 0.735          & 0.665              & 0.650           & 0.565          & 0.595              & 0.580          & 0.685          & 0.500             & 0.585           & 0.620              & 0.560          \\
        \midrule
        \textbf{Gemini 2 Flash}                        & 0.635            & 0.615            & \textbf{0.715}    & 0.650            & 0.765          & 0.635              & {\ul 0.740}     & \textbf{0.620} & 0.670              & \textbf{0.610} & 0.730          & \textbf{0.600}    & 0.550           & 0.635              & 0.510          \\
        \textbf{Gemini 2 Flash} (\textit{+POI name})   & 0.635            & 0.596            & {\ul 0.695}       & 0.640            & \textbf{0.790} & 0.630              & {\ul 0.740}     & {\ul 0.605}    & {\ul 0.675}        & \textbf{0.610} & \textbf{0.745} & {\ul 0.595}       & 0.540           & 0.645              & 0.490          \\
        \textbf{Claude Haiku 4.5}                      & 0.540            & 0.558            & 0.580             & 0.535            & 0.515          & 0.560              & 0.670           & 0.545          & 0.595              & 0.570          & 0.660          & 0.565             & 0.565           & 0.570              & 0.535          \\
        \textbf{Claude Haiku 4.5} (\textit{+POI name}) & 0.590            & 0.596            & 0.635             & 0.570            & 0.475          & 0.600              & 0.700           & 0.565          & 0.600              & 0.510          & 0.660          & 0.560             & 0.575           & 0.545              & {\ul 0.585}    \\
        \textbf{GPT-4o Mini}                           & 0.625            & 0.538            & 0.610             & 0.610            & 0.430          & 0.635              & \textbf{0.745}  & 0.600          & 0.645              & {\ul 0.590}    & 0.645          & 0.565             & 0.545           & 0.600              & 0.495          \\
        \textbf{GPT-4.1 Mini}                          & 0.585            & 0.673            & 0.615             & 0.600            & 0.690          & 0.585              & \textbf{0.745}  & 0.595          & 0.605              & 0.575          & 0.700          & 0.565             & 0.515           & 0.620              & 0.550          \\
        \textbf{GPT-5 Nano}                            & 0.570            & 0.635            & 0.535             & 0.530            & 0.470          & 0.500              & 0.635           & 0.580          & 0.560              & 0.565          & 0.680          & 0.465             & 0.440           & 0.520              & 0.580          \\
        \bottomrule
    \end{tabular}
\end{table*}

To examine temporal changes in mobility patterns, we split the test sets into two periods (2012-2013 and 2017-2018) and evaluated the same four LLMs with LLM-Move \cite{feng2024move}, which was the strongest-performing approach in our zero-shot experiments. As shown in Table~\ref{tab:longitudinal-zero-shot-results}, zero-shot accuracy generally declined in the 2017-2018 period, except for Jakarta and Tangerang, indicating that user trajectories in later years tend to be more challenging to predict. Performance trends varied across cities, highlighting temporal differences in mobility patterns that impact downstream tasks. Across all models, Gemini 2.0 Flash consistently achieved the highest accuracy, demonstrating robust zero-shot capabilities across cities and time.

\subsection{Spatiotemporal Classification and Reasoning}

This benchmark assesses whether LLMs can be leveraged for spatiotemporal trajectory classification by providing them with contextual information about a user's behavior. The task evaluates the model's ability to capture variations in travel patterns across different cities, given the sequence of POI check-ins as input, and without any additional fine-tuning. Through this setup, we aim to understand how effectively LLMs can reason over spatiotemporal and behavioral cues in user trajectories. Appendix~\ref{sec:suppl-st-day-zs} provides details on LLM parameters.

\subsubsection{Problem Formulation}

Borrowing the notation used in Section~\ref{sec:suppl-poi-rec}, we formulate this task as follows.
Given a current contextual trajectory $T'_u(t)$ of user $u$ up to time $t$, the goal of spatiotemporal trajectory classification is to predict a property $y$ of the trajectory. In this study, we focus on \textbf{weekday/weekend classification}, where $y \in \{\text{weekday}, \text{weekend}\}$. Formally, the LLM serves as a classification function:
\[
    f: T'_u(t) \rightarrow \hat{y}
\]
where $\hat{y}$ denotes the predicted class label for the trajectory. The model is evaluated based on its accuracy in correctly classifying trajectories according to this property.

\subsubsection{Experimental Setup}

This task involves trajectory property classification, specifically predicting whether the final check-in occurs on a weekday or a weekend. Each trajectory is converted into a textual prompt incorporating spatial (city), temporal (check-in time-of-day), categorical contexts (POI category), and semantic contexts (POI venue name). Adapting the prompt design from LLM-Mob \cite{wang2023would}, we instructed the LLM to first reason before making a prediction. This approach allows us to evaluate both classification accuracy and the spatiotemporal reasoning capabilities of LLMs, in line with recent work on spatiotemporal reasoning using LLMs \cite{quan2025benchmarkingspatiotemporalreasoningllms}. Whereas prior approaches rely on models that encode trajectories \cite{NAYAK2023100595}, our method directly leverages the LLM's ability to process contextual information in natural language. We evaluated five closed-source LLM APIs: Gemini 2.0 Flash \cite{geminiteam2024geminifamilyhighlycapable}, Claude Haiku 4.5, GPT-4o Mini, GPT-4.1 Mini \cite{openai2024gpt4technicalreport,openai2024gpt4ocard}, and GPT-5 Nano \cite{singh2025openai}, and used Accuracy as our primary metric.

\subsubsection{Baseline: Bag-of-POI Categories}

As a baseline, we also evaluated classical machine learning classifiers: Logistic Regression, Random Forest, and XGBoost \cite{10.1145/2939672.2939785}. For feature extraction, instead of textual prompting, we propose to simply transform each trajectory into a semantic sequence of POI categories. These sequences are then vectorized using either Bag-of-Words (BoW) or TF-IDF, treating POI categories as "words" and historical user trajectories as a corpus. Unlike the LLMs, these classical models are trained in a fully supervised manner and are very lightweight. 

\subsubsection{Main Results}

As shown in Table~\ref{tab:st-day-results}, Logistic Regression with TF-IDF achieves the highest mean accuracy (0.658) across the 15 cities, outperforming modern LLMs. The best-performing LLM (Gemini 2 Flash) achieves a mean accuracy of 0.645, performing slightly below classical ML baselines. This suggests that simple frequency-based representations of categorical features remain highly competitive for this task. Interestingly, the GPT series generally performs worse, even for models that are more recent than Gemini 2 Flash. For instance, GPT-5 Nano records the lowest mean accuracy, despite being designed to support more advanced reasoning capabilities. Moreover, adding semantic context does not consistently improve performance, and in some cases, it even leads to a decline, potentially due to variability in the names that do not always correlate with the temporal patterns being classified. Overall, these results highlight current limitations of LLMs in capturing spatiotemporal patterns from trajectory data alone. The stronger performance of classical baselines further indicates that, in a zero-shot setting, LLMs still struggle to reason over spatiotemporal patterns as effectively as statistical methods capture categorical regularities. These findings also suggest that fine-tuning on spatiotemporal data may be necessary for LLMs to surpass classical approaches.


\section{Conclusion}

\paragraph{Conclusion}
In this paper, we presented Massive-STEPS to address longstanding limitations in POI trajectory modeling, particularly the reliance on older, geographically saturated, and non-reproducible check-in datasets. Massive-STEPS offers a large-scale, semantically enriched POI check-in dataset spanning 15 cities across diverse regions and two time periods, supporting single-city, cross-city, and longitudinal studies. The dataset includes rich semantic information about a POI check-in record, including its name, address, category, and geographic coordinates. We also provide comprehensive benchmarking results for supervised POI recommendation, zero-shot POI recommendation, and spatiotemporal classification, demonstrating the dataset's versatility across models and tasks. Across these tasks, GNNs performed best for supervised POI recommendation, zero-shot LLMs were competitive in some cities, and classical ML models remained strong for spatiotemporal classification. By releasing our dataset and benchmark pipeline publicly, we aim to advance open, reproducible, and geographically inclusive research in POI modeling and human mobility studies.

\paragraph{Limitations}
\label{sec:limitations}

Firstly, Massive-STEPS is derived from Semantic Trails, thus inheriting its biases, which may propagate through downstream tasks. Additionally, the dataset is sparse in several cities and does not include African cities; we leave the latter for future work. Secondly, Massive-STEPS focuses solely on trajectories and POI metadata, without including user demographic or social information due to privacy considerations. This restricts its applicability for personalized or socially-aware POI recommendation tasks. Thirdly, while our benchmarking covers a wide range of models and cities to emphasize replicability and breadth, we did not perform extensive hyperparameter tuning, which may affect the peak performance of the models. Finally, although Massive-STEPS does not reflect present-day mobility patterns, it was designed to provide an alternative to older datasets and to help bridge the gap toward newer, open, and extensible POI benchmarks.



\begin{acks}
    We would like to thank the support of the National Computational Infrastructure and the ARC Centre of Excellence for Automated Decision Making and Society (CE200100005). Computational facilities were provided by the School of Computer Science and Engineering at UNSW Sydney through the Wolfpack computational cluster. Additionally, we express our gratitude to the NVIDIA Academic Grant Program for providing access to A100 GPUs on Saturn Cloud. We further thank the Google Cloud Research Credits program for API access to Gemini models.
\end{acks}

\bibliographystyle{ACM-Reference-Format}
\bibliography{sample-sigconf}

\clearpage

\appendix

\section{POI Recommendation: Details}
\label{sec:suppl-poi-rec}


For training and evaluation, we used the LibCity\footnote{\url{https://github.com/libcity/bigscity-libcity-datasets/}} library \cite{10.1145/3474717.3483923}, which provides implementations of classical baselines including FPMC \cite{10.1145/1772690.1772773}, RNN  \cite{10.1145/3474717.3483923}, LSTPM \cite{Sun_Qian_Chen_Liang_Nguyen_Yin_2020}, and DeepMove \cite{10.1145/3178876.3186058}. The training hyperparameters are listed in Table~\ref{tab:hypers-baseline} and, unless otherwise noted, follow the default configurations provided by LibCity. For Flashback\footnote{\url{https://github.com/exascaleinfolab/flashback_code/}} \cite{10.5555/3491440.3491742}, Graph-Flashback\footnote{\url{https://github.com/kevin-xuan/Graph-Flashback}} \cite{10.1145/3534678.3539383}, LoTNext\footnote{\url{https://github.com/Yukayo/LoTNext}} \cite{10.5555/3737916.3739652}, GETNext\footnote{\url{https://github.com/songyangme/GETNext}} \cite{10.1145/3477495.3531983}, and STHGCN\footnote{\url{https://github.com/alipay/Spatio-Temporal-Hypergraph-Model}} \cite{10.1145/3539618.3591770}, we adapted the source code released by the respective authors. Due to variations in dataset sizes and training costs across cities, we applied different hyperparameters for some cities, as detailed in Table~\ref{tab:hypers-gnns}. For UniMove\footnote{\url{https://github.com/tsinghua-fib-lab/unimove}} \cite{han2025unimoveunifiedmodelmulticity}, we modified their original source code for Massive-STEPS. For location features, we used Schema.org's 162 list of categories as a categorical feature and the administrative region as the grid area for POI category distribution. Hyperparameters are listed in Table~\ref{tab:hypers-baseline} and, unless otherwise noted, follow the default values. We report the full results of our supervised POI recommendation baselines in Table~\ref{tab:full-results-baseline-1} and \ref{tab:full-results-baseline-2}, using three evaluation metrics: Acc@1, Acc@5, and NDCG@5. All modified code implementations are available as submodules in our main dataset repository. Experiments were conducted using NVIDIA L4, L40S, and H100 GPUs.

\begin{table}[htbp]
    \centering
    \small
    \caption{\textbf{Hyperparameters for Markov-based methods, recurrent networks, and UniMove}.}
    \label{tab:hypers-baseline}
    \setlength{\tabcolsep}{1.5pt}
    \begin{tabular}{lccccc}
        \toprule
        \textbf{Hyperparameter} & \textbf{FPMC} & \textbf{RNN} & \textbf{LSTPM} & \textbf{DeepMove} & \textbf{UniMove} \\
        \midrule
        Batch Size              & 20            & 20           & 20             & 20                & 4                \\
        Learning Rate           & 5e-4          & 1e-3         & 1e-4           & 1e-3              & 3e-4             \\
        Max Epoch               & 1             & 30           & 40             & 30                & 50               \\
        Location Embedding Size & 64            & 500          & 500            & 500               & \{256, 128\}     \\
        Hidden Embedding Size   & N/A           & 500          & 500            & 500               & 512              \\
        Dropout                 & N/A           & 0.3          & 0.8            & 0.5               & N/A              \\
        \bottomrule
    \end{tabular}
\end{table}

\begin{table}[htbp]
    \centering
    \tiny
    \caption{\textbf{Hyperparameters for Transformer-based graph neural networks}.}
    \label{tab:hypers-gnns}
    \setlength{\tabcolsep}{2.5pt}
    \begin{tabular}{llccc}
        \toprule
        \textbf{Model}                    & \textbf{Cities}                                                & \textbf{BS} & \textbf{LR} & \textbf{Epochs} \\ \midrule
        \textbf{Flashback},               & Beijing                                                        & 10          & 1e-2        & 100             \\
        \textbf{Graph-Flashback},         & Bandung, Istanbul, Jakarta, KC, Melbourne, Moscow, NY          & 100         & 1e-2        & 100             \\
        \textbf{LoTNext}                  & Palembang, PJ, SP, Shanghai, Sydney, Tangerang, Tokyo          & 100         & 1e-2        & 100             \\
        \midrule
        \multirow{3}{*}{\textbf{GETNext}} & Beijing, Melbourne, Moscow, Palembang, Shanghai, Sydney, Tokyo & 16          & 1e-3        & 200             \\
                                          & Bandung, Istanbul, KC, NY, PJ, SP, Tangerang                   & 16          & 1e-4        & 20              \\
                                          & Jakarta                                                        & 16          & 5e-5        & 20              \\ \midrule
        \multirow{2}{*}{\textbf{STHGCN}}  & Beijing, Melbourne, Palembang, Shanghai, Sydney, Tokyo         & 16          & 1e-4        & 20              \\
                                          & Bandung, Istanbul, Jakarta, KC, Moscow, NY, PJ, SP, Tangerang  & 64          & 1e-4        & 20              \\
        \bottomrule
    \end{tabular}
\end{table}

\begin{table*}[htbp]
    \centering
    \tiny
    \caption{\textbf{Performance of supervised POI recommendation baselines across} Bandung, Beijing, Istanbul, Jakarta, Kuwait City, Melbourne, Moscow, New York.}
    \label{tab:full-results-baseline-1}
    \setlength{\tabcolsep}{3pt}
    \begin{tabular}{lccc|ccc|ccc|ccc|ccc|ccc|ccc|ccc}
        \toprule
        \multirow{2}{*}{\textbf{Model}} & \multicolumn{3}{c}{\textbf{Bandung}} & \multicolumn{3}{c}{\textbf{Beijing}} & \multicolumn{3}{c}{\textbf{Istanbul}} & \multicolumn{3}{c}{\textbf{Jakarta}} & \multicolumn{3}{c}{\textbf{Kuwait City}} & \multicolumn{3}{c}{\textbf{Melbourne}} & \multicolumn{3}{c}{\textbf{Moscow}} & \multicolumn{3}{c}{\textbf{New York}}                                                                                                                                                                                                                                                 \\
        \cmidrule{2-25}
                                        & \textbf{A@1}                         & \textbf{A@5}                         & \textbf{N@5}                          & \textbf{A@1}                         & \textbf{A@5}                             & \textbf{N@5}                           & \textbf{A@1}                        & \textbf{A@5}                          & \textbf{N@5} & \textbf{A@1} & \textbf{A@5} & \textbf{N@5} & \textbf{A@1} & \textbf{A@5} & \textbf{N@5} & \textbf{A@1} & \textbf{A@5} & \textbf{N@5} & \textbf{A@1} & \textbf{A@5} & \textbf{N@5} & \textbf{A@1} & \textbf{A@5} & \textbf{N@5} \\
        \midrule
        \textbf{FPMC}                   & 0.048                                & 0.118                                & 0.083                                 & 0.000                                & 0.021                                    & 0.009                                  & 0.026                               & 0.074                                 & 0.050        & 0.029        & 0.085        & 0.058        & 0.021        & 0.089        & 0.054        & 0.062        & 0.147        & 0.107        & 0.059        & 0.129        & 0.094        & 0.032        & 0.090        & 0.061        \\
        \midrule
        \textbf{RNN}                    & 0.062                                & 0.135                                & 0.099                                 & 0.085                                & 0.183                                    & 0.134                                  & 0.077                               & 0.178                                 & 0.130        & 0.049        & 0.115        & 0.083        & 0.087        & 0.203        & 0.146        & 0.059        & 0.105        & 0.083        & 0.075        & 0.164        & 0.122        & 0.061        & 0.119        & 0.092        \\
        \textbf{LSTPM}                  & 0.110                                & 0.241                                & 0.179                                 & 0.127                                & 0.211                                    & 0.169                                  & 0.142                               & 0.286                                 & 0.217        & 0.099        & 0.210        & 0.157        & 0.180        & 0.362        & 0.275        & 0.091        & 0.204        & 0.150        & 0.151        & 0.300        & 0.229        & 0.099        & 0.206        & 0.155        \\
        \textbf{DeepMove}               & 0.107                                & 0.232                                & 0.172                                 & 0.106                                & 0.261                                    & 0.190                                  & 0.150                               & 0.298                                 & 0.228        & 0.103        & 0.212        & 0.160        & 0.179        & 0.360        & 0.274        & 0.083        & 0.179        & 0.134        & 0.143        & 0.283        & 0.217        & 0.097        & 0.195        & 0.149        \\
        \textbf{Flashback}              & 0.103                                & 0.202                                & 0.155                                 & 0.278                                & 0.435                                    & 0.359                                  & 0.137                               & 0.257                                 & 0.200        & 0.113        & 0.214        & 0.166        & 0.150        & 0.298        & 0.227        & 0.079        & 0.146        & 0.115        & 0.121        & 0.234        & 0.180        & 0.086        & 0.156        & 0.123        \\
        \textbf{Graph-Flashback}        & 0.091                                & 0.197                                & 0.146                                 & 0.304                                & 0.426                                    & 0.373                                  & 0.137                               & 0.261                                 & 0.202        & 0.072        & 0.167        & 0.121        & 0.115        & 0.245        & 0.181        & 0.093        & 0.188        & 0.143        & 0.125        & 0.243        & 0.188        & 0.059        & 0.125        & 0.093        \\
        \textbf{LoTNext}                & 0.076                                & 0.169                                & 0.125                                 & 0.243                                & 0.391                                    & 0.320                                  & 0.105                               & 0.227                                 & 0.168        & 0.044        & 0.110        & 0.078        & 0.132        & 0.280        & 0.209        & 0.085        & 0.180        & 0.133        & 0.089        & 0.196        & 0.145        & 0.061        & 0.133        & 0.099        \\
        \midrule
        \textbf{GETNext}                & 0.179                                & 0.306                                & 0.247                                 & 0.433                                & 0.527                                    & 0.486                                  & 0.146                               & 0.268                                 & 0.210        & 0.155        & 0.257        & 0.209        & 0.175        & 0.322        & 0.251        & 0.100        & 0.250        & 0.179        & 0.175        & 0.335        & 0.260        & 0.134        & 0.263        & 0.202        \\
        \textbf{STHGCN}                 & 0.219                                & 0.375                                & 0.302                                 & 0.453                                & 0.640                                    & 0.552                                  & 0.241                               & 0.385                                 & 0.318        & 0.197        & 0.334        & 0.270        & 0.225        & 0.394        & 0.314        & 0.168        & 0.318        & 0.247        & 0.223        & 0.382        & 0.308        & 0.146        & 0.259        & 0.207        \\
        \midrule
        \textbf{GNPR-SID}               & 0.126                                & 0.145                                & 0.136                                 & 0.139                                & 0.148                                    & 0.143                                  & 0.115                               & 0.137                                 & 0.127        & 0.102        & 0.112        & 0.107        & 0.162        & 0.192        & 0.178        & 0.049        & 0.060        & 0.054        & 0.116        & 0.138        & 0.128        & 0.074        & 0.085        & 0.080        \\
        \textbf{UniMove}                & 0.007                                & 0.060                                & 0.033                                 & 0.036                                & 0.205                                    & 0.128                                  & 0.015                               & 0.061                                 & 0.038        & 0.004        & 0.036        & 0.020        & 0.023        & 0.120        & 0.073        & 0.008        & 0.066        & 0.037        & 0.009        & 0.051        & 0.030        & 0.004        & 0.028        & 0.016        \\
        \bottomrule
    \end{tabular}
\end{table*}

\begin{table*}[htbp]
    \centering
    \tiny
    \caption{\textbf{Performance of supervised POI recommendation baselines across} Palembang, Petaling Jaya, São Paulo, Shanghai, Sydney, Tangerang, Tokyo.}
    \label{tab:full-results-baseline-2}
    \setlength{\tabcolsep}{3pt}
    \begin{tabular}{lccc|ccc|ccc|ccc|ccc|ccc|ccc}
        \toprule
        \multirow{2}{*}{\textbf{Model}} & \multicolumn{3}{c}{\textbf{Palembang}} & \multicolumn{3}{c}{\textbf{Petaling Jaya}} & \multicolumn{3}{c}{\textbf{São Paulo}} & \multicolumn{3}{c}{\textbf{Shanghai}} & \multicolumn{3}{c}{\textbf{Sydney}} & \multicolumn{3}{c}{\textbf{Tangerang}} & \multicolumn{3}{c}{\textbf{Tokyo}}                                                                                                                                                                                                                   \\
        \cmidrule{2-22}
                                        & \textbf{A@1}                           & \textbf{A@5}                               & \textbf{N@5}                           & \textbf{A@1}                          & \textbf{A@5}                        & \textbf{N@5}                           & \textbf{A@1}                       & \textbf{A@5} & \textbf{N@5} & \textbf{A@1} & \textbf{A@5} & \textbf{N@5} & \textbf{A@1} & \textbf{A@5} & \textbf{N@5} & \textbf{A@1} & \textbf{A@5} & \textbf{N@5} & \textbf{A@1} & \textbf{A@5} & \textbf{N@5} \\
        \midrule
        \textbf{FPMC}                   & 0.102                                  & 0.169                                      & 0.136                                  & 0.026                                 & 0.084                               & 0.057                                  & 0.030                              & 0.079        & 0.055        & 0.084        & 0.154        & 0.120        & 0.075        & 0.180        & 0.131        & 0.104        & 0.220        & 0.166        & 0.176        & 0.291        & 0.239        \\
        \midrule
        \textbf{RNN}                    & 0.049                                  & 0.121                                      & 0.085                                  & 0.064                                 & 0.148                               & 0.107                                  & 0.097                              & 0.191        & 0.147        & 0.055        & 0.120        & 0.090        & 0.080        & 0.164        & 0.125        & 0.087        & 0.179        & 0.135        & 0.133        & 0.254        & 0.197        \\
        \textbf{LSTPM}                  & 0.114                                  & 0.230                                      & 0.175                                  & 0.099                                 & 0.222                               & 0.163                                  & 0.158                              & 0.319        & 0.243        & 0.099        & 0.195        & 0.149        & 0.141        & 0.265        & 0.206        & 0.154        & 0.309        & 0.237        & 0.225        & 0.394        & 0.315        \\
        \textbf{DeepMove}               & 0.084                                  & 0.191                                      & 0.139                                  & 0.112                                 & 0.234                               & 0.175                                  & 0.160                              & 0.310        & 0.240        & 0.085        & 0.168        & 0.128        & 0.129        & 0.240        & 0.188        & 0.145        & 0.285        & 0.219        & 0.201        & 0.362        & 0.288        \\
        \textbf{Flashback}              & 0.135                                  & 0.256                                      & 0.198                                  & 0.072                                 & 0.162                               & 0.119                                  & 0.159                              & 0.280        & 0.222        & 0.096        & 0.162        & 0.132        & 0.127        & 0.216        & 0.174        & 0.185        & 0.330        & 0.262        & 0.170        & 0.298        & 0.237        \\
        \textbf{Graph-Flashback}        & 0.148                                  & 0.264                                      & 0.208                                  & 0.072                                 & 0.164                               & 0.119                                  & 0.096                              & 0.211        & 0.156        & 0.111        & 0.187        & 0.151        & 0.135        & 0.234        & 0.188        & 0.200        & 0.373        & 0.292        & 0.180        & 0.329        & 0.259        \\
        \textbf{LoTNext}                & 0.140                                  & 0.271                                      & 0.208                                  & 0.055                                 & 0.135                               & 0.097                                  & 0.105                              & 0.222        & 0.166        & 0.093        & 0.179        & 0.139        & 0.123        & 0.225        & 0.176        & 0.126        & 0.249        & 0.190        & 0.191        & 0.315        & 0.258        \\
        \midrule
        \textbf{GETNext}                & 0.158                                  & 0.313                                      & 0.239                                  & 0.139                                 & 0.254                               & 0.200                                  & 0.202                              & 0.360        & 0.286        & 0.115        & 0.230        & 0.177        & 0.181        & 0.347        & 0.266        & 0.224        & 0.372        & 0.302        & 0.180        & 0.361        & 0.275        \\
        \textbf{STHGCN}                 & 0.246                                  & 0.427                                      & 0.341                                  & 0.174                                 & 0.301                               & 0.241                                  & 0.250                              & 0.425        & 0.344        & 0.193        & 0.329        & 0.264        & 0.227        & 0.378        & 0.307        & 0.293        & 0.492        & 0.400        & 0.250        & 0.432        & 0.350        \\
        \midrule
        \textbf{GNPR-SID}               & 0.094                                  & 0.111                                      & 0.103                                  & 0.090                                 & 0.105                               & 0.098                                  & 0.165                              & 0.186        & 0.176        & 0.060        & 0.071        & 0.067        & 0.105        & 0.118        & 0.112        & 0.165        & 0.188        & 0.177        & 0.114        & 0.136        & 0.126        \\
        \textbf{UniMove}                & 0.009                                  & 0.060                                      & 0.035                                  & 0.008                                 & 0.058                               & 0.034                                  & 0.002                              & 0.018        & 0.009        & 0.000        & 0.055        & 0.029        & 0.015        & 0.102        & 0.059        & 0.001        & 0.055        & 0.029        & 0.032        & 0.109        & 0.072        \\
        \bottomrule
    \end{tabular}
\end{table*}


\section{Zero-shot POI Recommendation: Details}
\label{sec:suppl-poi-rec-zs}

We adopted the AgentMove\footnote{\url{https://github.com/tsinghua-fib-lab/agentmove/}} library \cite{feng-etal-2025-agentmove}, which provides implementations of LLM-Mob \cite{wang2023would}, LLM-ZS \cite{beneduce2024large}, and LLM-Move \cite{feng2024move}. The preprocessing steps used by AgentMove are as follows. First, we selected 200 random users from the test set and sampled one random trajectory for each user. This trajectory serves as the \textbf{context stays}, representing the current trajectory to be predicted. The \textbf{historical stays} are composed of the most recent 15 trajectories from the same user, drawn from the training set. Each check-in is described by four attributes: the hour (in 12-hour format), the day of the week, the POI ID, and the POI category name. Second, the LLMs are set to return outputs in JSON format, generating the top 5 predicted POI IDs along with an explanation of their reasoning. Following the AgentMove setup and to ensure replicability, we set the generation parameters as follows: a temperature of 0.0, a maximum output length of 1000 tokens, and an input context window capped at 2000 tokens. Following its default configuration, LLM-Move’s candidate set is set to the 100 nearest POIs. We provide the full results of our zero-shot POI recommendation results in Table~\ref{tab:full-results-zs-1} and \ref{tab:full-results-zs-2}, providing three metrics: Acc@1, Acc@5, and NDCG@5.

\begin{table*}[htbp]
    \centering
    \tiny
    \caption{\textbf{Performance of zero-shot POI recommendation baselines across} Bandung, Beijing, Istanbul, Jakarta, Kuwait City, Melbourne, Moscow, New York.}
    \label{tab:full-results-zs-1}
    \setlength{\tabcolsep}{3pt}
    \begin{tabular}{llccc|ccc|ccc|ccc|ccc|ccc|ccc|ccc}
        \toprule
        \multirow{2}{*}{\textbf{Method}}   & \multirow{2}{*}{\textbf{Model}} & \multicolumn{3}{c}{\textbf{Bandung}} & \multicolumn{3}{c}{\textbf{Beijing}} & \multicolumn{3}{c}{\textbf{Istanbul}} & \multicolumn{3}{c}{\textbf{Jakarta}} & \multicolumn{3}{c}{\textbf{Kuwait City}} & \multicolumn{3}{c}{\textbf{Melbourne}} & \multicolumn{3}{c}{\textbf{Moscow}} & \multicolumn{3}{c}{\textbf{New York}}                                                                                                                                                                                                                                                 \\
        \cmidrule{3-26}
                                           &                                 & \textbf{A@1}                         & \textbf{A@5}                         & \textbf{N@5}                          & \textbf{A@1}                         & \textbf{A@5}                             & \textbf{N@5}                           & \textbf{A@1}                        & \textbf{A@5}                          & \textbf{N@5} & \textbf{A@1} & \textbf{A@5} & \textbf{N@5} & \textbf{A@1} & \textbf{A@5} & \textbf{N@5} & \textbf{A@1} & \textbf{A@5} & \textbf{N@5} & \textbf{A@1} & \textbf{A@5} & \textbf{N@5} & \textbf{A@1} & \textbf{A@5} & \textbf{N@5} \\
        \midrule
        \multirow{4}{*}{\textbf{LLM-Mob}}  & \textbf{Gemini 2 Flash}         & 0.105                                & 0.170                                & 0.139                                 & 0.115                                & 0.308                                    & 0.226                                  & 0.080                               & 0.225                                 & 0.160        & 0.100        & 0.245        & 0.174        & 0.095        & 0.270        & 0.185        & 0.060        & 0.150        & 0.108        & 0.130        & 0.245        & 0.187        & 0.095        & 0.175        & 0.136        \\
                                           & \textbf{Claude Haiku 4.5}       & 0.115                                & 0.190                                & 0.150                                 & 0.115                                & 0.404                                    & 0.268                                  & 0.085                               & 0.265                                 & 0.182        & 0.135        & 0.260        & 0.201        & 0.110        & 0.270        & 0.190        & 0.070        & 0.155        & 0.113        & 0.130        & 0.280        & 0.203        & 0.075        & 0.180        & 0.128        \\
                                           & \textbf{Qwen 2.5 7B}            & 0.060                                & 0.155                                & 0.111                                 & 0.058                                & 0.385                                    & 0.218                                  & 0.035                               & 0.240                                 & 0.148        & 0.105        & 0.245        & 0.179        & 0.080        & 0.220        & 0.155        & 0.030        & 0.130        & 0.083        & 0.090        & 0.270        & 0.185        & 0.070        & 0.185        & 0.131        \\
                                           & \textbf{Llama 3.1 8B}           & 0.010                                & 0.100                                & 0.055                                 & 0.000                                & 0.000                                    & 0.000                                  & 0.020                               & 0.110                                 & 0.065        & 0.055        & 0.150        & 0.104        & 0.030        & 0.100        & 0.066        & 0.010        & 0.065        & 0.040        & 0.030        & 0.100        & 0.068        & 0.025        & 0.090        & 0.061        \\
                                           & \textbf{Gemma 2 9B}             & 0.070                                & 0.175                                & 0.126                                 & 0.115                                & 0.288                                    & 0.206                                  & 0.075                               & 0.200                                 & 0.146        & 0.105        & 0.240        & 0.178        & 0.080        & 0.210        & 0.150        & 0.055        & 0.150        & 0.108        & 0.100        & 0.240        & 0.176        & 0.070        & 0.175        & 0.124        \\
        \midrule
        \multirow{4}{*}{\textbf{LLM-ZS}}   & \textbf{Gemini 2 Flash}         & 0.095                                & 0.195                                & 0.147                                 & 0.058                                & 0.385                                    & 0.246                                  & 0.090                               & 0.235                                 & 0.166        & 0.110        & 0.250        & 0.188        & 0.080        & 0.245        & 0.167        & 0.065        & 0.160        & 0.115        & 0.125        & 0.300        & 0.217        & 0.080        & 0.170        & 0.129        \\
                                           & \textbf{Claude Haiku 4.5}       & 0.115                                & 0.195                                & 0.156                                 & 0.115                                & 0.385                                    & 0.273                                  & 0.100                               & 0.250                                 & 0.178        & 0.145        & 0.265        & 0.206        & 0.130        & 0.280        & 0.207        & 0.070        & 0.140        & 0.107        & 0.130        & 0.280        & 0.206        & 0.090        & 0.190        & 0.141        \\
                                           & \textbf{Qwen 2.5 7B}            & 0.055                                & 0.185                                & 0.126                                 & 0.038                                & 0.404                                    & 0.237                                  & 0.040                               & 0.235                                 & 0.141        & 0.065        & 0.250        & 0.161        & 0.050        & 0.220        & 0.140        & 0.040        & 0.155        & 0.100        & 0.080        & 0.260        & 0.176        & 0.050        & 0.180        & 0.116        \\
                                           & \textbf{Llama 3.1 8B}           & 0.045                                & 0.210                                & 0.131                                 & 0.077                                & 0.346                                    & 0.221                                  & 0.040                               & 0.225                                 & 0.137        & 0.045        & 0.200        & 0.126        & 0.060        & 0.210        & 0.137        & 0.040        & 0.155        & 0.101        & 0.080        & 0.270        & 0.183        & 0.055        & 0.160        & 0.111        \\
                                           & \textbf{Gemma 2 9B}             & 0.065                                & 0.185                                & 0.130                                 & 0.096                                & 0.308                                    & 0.217                                  & 0.045                               & 0.225                                 & 0.141        & 0.105        & 0.250        & 0.180        & 0.070        & 0.230        & 0.153        & 0.050        & 0.140        & 0.100        & 0.080        & 0.290        & 0.194        & 0.075        & 0.185        & 0.129        \\
        \midrule
        \multirow{4}{*}{\textbf{LLM-Move}} & \textbf{Gemini 2 Flash}         & 0.225                                & 0.350                                & 0.289                                 & 0.096                                & 0.346                                    & 0.218                                  & 0.205                               & 0.385                                 & 0.289        & 0.295        & 0.405        & 0.350        & 0.220        & 0.380        & 0.295        & 0.225        & 0.325        & 0.275        & 0.220        & 0.400        & 0.316        & 0.235        & 0.415        & 0.325        \\
                                           & \textbf{Claude Haiku 4.5}       & 0.205                                & 0.270                                & 0.240                                 & 0.154                                & 0.288                                    & 0.227                                  & 0.195                               & 0.395                                 & 0.295        & 0.260        & 0.315        & 0.287        & 0.185        & 0.375        & 0.283        & 0.175        & 0.330        & 0.249        & 0.230        & 0.425        & 0.330        & 0.170        & 0.305        & 0.241        \\
                                           & \textbf{Qwen 2.5 7B}            & 0.100                                & 0.155                                & 0.128                                 & 0.192                                & 0.346                                    & 0.280                                  & 0.175                               & 0.270                                 & 0.226        & 0.115        & 0.225        & 0.169        & 0.160        & 0.285        & 0.227        & 0.110        & 0.220        & 0.165        & 0.230        & 0.310        & 0.274        & 0.120        & 0.255        & 0.188        \\
                                           & \textbf{Llama 3.1 8B}           & 0.030                                & 0.035                                & 0.033                                 & 0.058                                & 0.135                                    & 0.100                                  & 0.015                               & 0.055                                 & 0.036        & 0.015        & 0.025        & 0.021        & 0.010        & 0.035        & 0.023        & 0.040        & 0.195        & 0.123        & 0.005        & 0.065        & 0.031        & 0.035        & 0.130        & 0.084        \\
                                           & \textbf{Gemma 2 9B}             & 0.175                                & 0.245                                & 0.213                                 & 0.096                                & 0.365                                    & 0.229                                  & 0.100                               & 0.200                                 & 0.155        & 0.235        & 0.290        & 0.266        & 0.120        & 0.275        & 0.202        & 0.115        & 0.275        & 0.199        & 0.110        & 0.245        & 0.185        & 0.115        & 0.245        & 0.183        \\
        \bottomrule
    \end{tabular}
\end{table*}

\begin{table*}[htbp]
    \centering
    \tiny
    \caption{\textbf{Performance of zero-shot POI recommendation baselines across} Palembang, Petaling Jaya, São Paulo, Shanghai, Sydney, Tangerang, Tokyo.}
    \label{tab:full-results-zs-2}
    \setlength{\tabcolsep}{3pt}
    \begin{tabular}{llccc|ccc|ccc|ccc|ccc|ccc|ccc}
        \toprule
        \multirow{2}{*}{\textbf{Method}}   & \multirow{2}{*}{\textbf{Model}} & \multicolumn{3}{c}{\textbf{Palembang}} & \multicolumn{3}{c}{\textbf{Petaling Jaya}} & \multicolumn{3}{c}{\textbf{São Paulo}} & \multicolumn{3}{c}{\textbf{Shanghai}} & \multicolumn{3}{c}{\textbf{Sydney}} & \multicolumn{3}{c}{\textbf{Tangerang}} & \multicolumn{3}{c}{\textbf{Tokyo}}                                                                                                                                                                                                                   \\
        \cmidrule{3-23}
                                           &                                 & \textbf{A@1}                           & \textbf{A@5}                               & \textbf{N@5}                           & \textbf{A@1}                          & \textbf{A@5}                        & \textbf{N@5}                           & \textbf{A@1}                       & \textbf{A@5} & \textbf{N@5} & \textbf{A@1} & \textbf{A@5} & \textbf{N@5} & \textbf{A@1} & \textbf{A@5} & \textbf{N@5} & \textbf{A@1} & \textbf{A@5} & \textbf{N@5} & \textbf{A@1} & \textbf{A@5} & \textbf{N@5} \\
        \midrule
        \multirow{4}{*}{\textbf{LLM-Mob}}  & \textbf{Gemini 2 Flash}         & 0.135                                  & 0.275                                      & 0.208                                  & 0.090                                 & 0.220                               & 0.160                                  & 0.130                              & 0.305        & 0.223        & 0.055        & 0.160        & 0.111        & 0.060        & 0.160        & 0.112        & 0.155        & 0.285        & 0.225        & 0.140        & 0.320        & 0.238        \\
                                           & \textbf{Claude Haiku 4.5}       & 0.135                                  & 0.285                                      & 0.214                                  & 0.090                                 & 0.210                               & 0.156                                  & 0.170                              & 0.330        & 0.251        & 0.070        & 0.170        & 0.122        & 0.075        & 0.165        & 0.119        & 0.155        & 0.320        & 0.244        & 0.180        & 0.400        & 0.298        \\
                                           & \textbf{Qwen 2.5 7B}            & 0.075                                  & 0.205                                      & 0.143                                  & 0.030                                 & 0.195                               & 0.116                                  & 0.090                              & 0.290        & 0.188        & 0.040        & 0.170        & 0.108        & 0.035        & 0.145        & 0.091        & 0.095        & 0.285        & 0.196        & 0.110        & 0.350        & 0.243        \\
                                           & \textbf{Llama 3.1 8B}           & 0.005                                  & 0.040                                      & 0.025                                  & 0.010                                 & 0.090                               & 0.050                                  & 0.030                              & 0.165        & 0.098        & 0.005        & 0.020        & 0.013        & 0.020        & 0.085        & 0.053        & 0.020        & 0.120        & 0.073        & 0.005        & 0.045        & 0.025        \\
                                           & \textbf{Gemma 2 9B}             & 0.095                                  & 0.240                                      & 0.171                                  & 0.055                                 & 0.185                               & 0.122                                  & 0.085                              & 0.230        & 0.162        & 0.050        & 0.150        & 0.104        & 0.030        & 0.130        & 0.086        & 0.145        & 0.270        & 0.209        & 0.145        & 0.345        & 0.255        \\
        \midrule
        \multirow{4}{*}{\textbf{LLM-ZS}}   & \textbf{Gemini 2 Flash}         & 0.130                                  & 0.260                                      & 0.196                                  & 0.110                                 & 0.210                               & 0.164                                  & 0.150                              & 0.315        & 0.235        & 0.065        & 0.160        & 0.113        & 0.060        & 0.155        & 0.111        & 0.145        & 0.310        & 0.234        & 0.160        & 0.380        & 0.278        \\
                                           & \textbf{Claude Haiku 4.5}       & 0.165                                  & 0.280                                      & 0.227                                  & 0.120                                 & 0.215                               & 0.174                                  & 0.170                              & 0.330        & 0.251        & 0.075        & 0.170        & 0.128        & 0.070        & 0.155        & 0.116        & 0.175        & 0.325        & 0.254        & 0.205        & 0.395        & 0.306        \\
                                           & \textbf{Qwen 2.5 7B}            & 0.050                                  & 0.215                                      & 0.135                                  & 0.045                                 & 0.175                               & 0.111                                  & 0.095                              & 0.290        & 0.198        & 0.045        & 0.155        & 0.103        & 0.045        & 0.170        & 0.109        & 0.100        & 0.315        & 0.215        & 0.120        & 0.365        & 0.257        \\
                                           & \textbf{Llama 3.1 8B}           & 0.070                                  & 0.240                                      & 0.154                                  & 0.030                                 & 0.205                               & 0.123                                  & 0.030                              & 0.280        & 0.159        & 0.060        & 0.165        & 0.116        & 0.040        & 0.185        & 0.110        & 0.080        & 0.255        & 0.173        & 0.110        & 0.415        & 0.269        \\
                                           & \textbf{Gemma 2 9B}             & 0.060                                  & 0.235                                      & 0.150                                  & 0.065                                 & 0.185                               & 0.126                                  & 0.075                              & 0.300        & 0.192        & 0.050        & 0.165        & 0.112        & 0.045        & 0.155        & 0.103        & 0.100        & 0.330        & 0.227        & 0.110        & 0.395        & 0.263        \\
        \midrule
        \multirow{4}{*}{\textbf{LLM-Move}} & \textbf{Gemini 2 Flash}         & 0.260                                  & 0.385                                      & 0.329                                  & 0.210                                 & 0.335                               & 0.273                                  & 0.285                              & 0.415        & 0.350        & 0.170        & 0.270        & 0.221        & 0.230        & 0.420        & 0.331        & 0.200        & 0.340        & 0.274        & 0.250        & 0.470        & 0.368        \\
                                           & \textbf{Claude Haiku 4.5}       & 0.175                                  & 0.215                                      & 0.195                                  & 0.165                                 & 0.245                               & 0.208                                  & 0.275                              & 0.375        & 0.325        & 0.100        & 0.175        & 0.141        & 0.235        & 0.415        & 0.325        & 0.220        & 0.330        & 0.275        & 0.255        & 0.415        & 0.340        \\
                                           & \textbf{Qwen 2.5 7B}            & 0.130                                  & 0.195                                      & 0.163                                  & 0.135                                 & 0.175                               & 0.155                                  & 0.155                              & 0.235        & 0.199        & 0.095        & 0.165        & 0.133        & 0.125        & 0.280        & 0.205        & 0.175        & 0.280        & 0.229        & 0.250        & 0.360        & 0.312        \\
                                           & \textbf{Llama 3.1 8B}           & 0.010                                  & 0.015                                      & 0.013                                  & 0.040                                 & 0.060                               & 0.049                                  & 0.045                              & 0.045        & 0.045        & 0.020        & 0.040        & 0.030        & 0.055        & 0.220        & 0.141        & 0.000        & 0.005        & 0.003        & 0.030        & 0.060        & 0.046        \\
                                           & \textbf{Gemma 2 9B}             & 0.210                                  & 0.270                                      & 0.240                                  & 0.175                                 & 0.235                               & 0.208                                  & 0.195                              & 0.300        & 0.252        & 0.105        & 0.150        & 0.128        & 0.125        & 0.370        & 0.254        & 0.125        & 0.250        & 0.193        & 0.130        & 0.305        & 0.225        \\
        \bottomrule
    \end{tabular}
\end{table*}

\section{Spatiotemporal Classification and Reasoning: Details}
\label{sec:suppl-st-day-zs}

We borrowed the experimental setup of AgentMove, similar to our zero-shot POI recommendation procedure. That is, we selected 200 random users from the test set and sampled one random trajectory for each user. This trajectory is then included in the test set. Each check-in is described by four attributes: the hour (in 12-hour format), the day of the week, the POI ID, and the POI category name.
LLMs are set to return outputs in a structured/JSON format, predicting whether the trajectory ended on a weekday or a weekend, along with an explanation of their reasoning. To ensure replicability, Gemini and GPT-4 models are set with the following generation parameters: a temperature of 0.0, a maximum output length of 1000 tokens, and an input context window capped at 2000 tokens. Due to API requirements, GPT-5 Nano uses the following generation parameters: a fixed temperature of 1.0, a maximum output length of 4096 tokens, low verbosity, and medium reasoning effort.

\end{document}